\pdfoutput=1

\documentclass[11pt]{article}

\usepackage{graphicx}

\usepackage[normalem]{ulem}
\useunder{\uline}{\ul}{}

\usepackage{multirow}
\usepackage{times}
\usepackage{latexsym}
\usepackage{graphicx}
\usepackage{algorithm}
\usepackage{algorithmic}
\usepackage{amssymb}
\usepackage{tikz,lipsum}
\usepackage{booktabs} 
\usepackage{amsmath,amsfonts}
\usepackage[export]{adjustbox}
\usepackage{tcolorbox}
\usepackage{enumitem}
\setlist{leftmargin=2mm}
\usepackage{pifont}
\usepackage{setspace}
\usepackage[export]{adjustbox}
\definecolor{mycolor}{HTML}{08872a}

\usepackage{float}

\usepackage[draft,textsize=footnotesize,textwidth=15mm]{todonotes}

\newcommand\acb[1]{\textcolor{blue}{#1}}

\usepackage{EMNLP2023}

\usepackage{times}
\usepackage{latexsym}

\usepackage[T1]{fontenc}

\usepackage[utf8]{inputenc}

\usepackage{microtype}

\usepackage{inconsolata}

\title{\textit{MemeGuard}: An LLM and VLM-based Framework for Advancing Content Moderation via Meme Intervention}

\author{Prince Jha$^{1}$,
Raghav Jain$^{1}$,
Konika Mandal$^{1}$,
Aman Chadha$^{2}$\thanks{\,\,\,Work does not relate to position at Amazon.}\,\,,
Sriparna Saha$^{1}$, \\
\and \textbf{Pushpak Bhattacharyya}$^{3}$\\
$^{1}$Department of Computer Science and Engineering, Indian Institute of Technology Patna\\
$^{2}$Amazon AI\\$^{3}$Department of Computer Science and Engineering, Indian Institute of Technology Bombay
}

\begin{document}
\maketitle
\begin{abstract}
In the digital world, memes present a unique challenge for content moderation due to their potential to spread harmful content. Although detection methods have improved, proactive solutions such as intervention are still limited, with current research focusing mostly on text-based content, neglecting the widespread influence of multimodal content like memes. Addressing this gap, we present \textit{MemeGuard}, a comprehensive framework leveraging Large Language Models (LLMs) and Visual Language Models (VLMs) for meme intervention. \textit{MemeGuard} harnesses a specially fine-tuned VLM, \textit{VLMeme}, for meme interpretation, and a multimodal knowledge selection and ranking mechanism (\textit{MKS}) for distilling relevant knowledge. This knowledge is then employed by a general-purpose LLM to generate contextually appropriate interventions. Another key contribution of this work is the \textit{\textbf{I}ntervening} \textit{\textbf{C}yberbullying in \textbf{M}ultimodal \textbf{M}emes (ICMM)} dataset, a high-quality, labeled dataset featuring toxic memes and their corresponding human-annotated interventions. We leverage \textit{ICMM} to test \textit{MemeGuard}, demonstrating its proficiency in generating relevant and effective responses to toxic memes.\footnote{Code and dataset are available at \url{https://github.com/Jhaprince/MemeGuard}}\\
 {\color{red} \textbf{Disclaimer}: \textit{This paper contains harmful content that may be disturbing to some readers.}}

\end{abstract}

\section{Introduction}

\begin{figure}[hbt]
	\centering
\vspace{-1mm} 
	\includegraphics[height = 4.5 cm, width = 7cm]{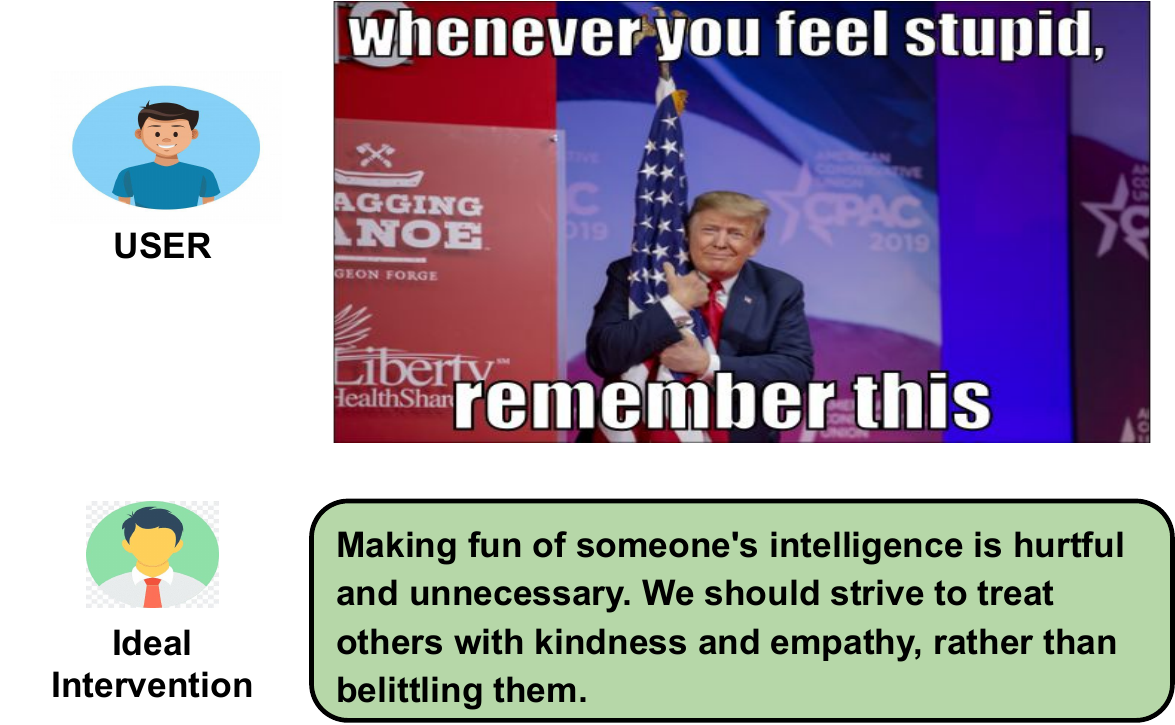}
 \vspace{-0.5mm}
	\caption {An instance of the meme intervention task.}
	\label{intro}
\vspace{-5mm} 
\end{figure}

In today's digital world, memes serve as a universal language for expression and engagement. However, as they become a powerful tool for rapid information dissemination, they are increasingly weaponized for cyberbullying and spreading toxic content, posing a challenge to existing content moderation systems. These systems struggle to decipher memes' nuanced meanings, typically reacting \cite{ohlheiser2016banned} rather than proactively mitigating the harm of offensive content.


\textls[-10]{Intervention represents a proactive approach to content moderation, going beyond simple detection to take preventive action against offensive content. Interventions aim to mitigate the harmful effects of toxic content and foster a more positive and respectful online discourse. However, existing intervention research has primarily been limited to text-based content such as hate speech \cite{qian2019benchmark}, and misinformation \cite{he2023reinforcement}. The excessive focus on text-based content overlooks the prevalence of multimodal content, which are major contributors to the content ecosystem in social media platforms. This exacerbates the potential for multimodal toxicity, enabling misuse of such mediums. \textit{A representative example of intervening in case of a cyberbullying meme is displayed in Figure \ref{intro}.}}\par
Large Language Models (LLMs) \cite{kojima2023large}, and Visual Language Models (VLMs) \cite{ghosh2024exploring} have shown remarkable capability in understanding and generating human-like text and multimedia content. This has led to their application in various tasks within the domain of content moderation, but predominantly for detection purposes. Their ability to understand nuanced language and visual cues has been leveraged to detect toxic or harmful content, both in text \cite{elsherief-etal-2021-latent,  DBLP:conf/emnlp/MaityJJ0B23, 10494986} and multimodal formats \cite{10.1145/3477495.3531925, DBLP:conf/ijcnn/JainMJS23, DBLP:conf/eacl/JhaMJVSB24}. Some attempts have also been made to use these models for intervention generation tasks, but these efforts have been largely restricted to text-based content \cite{qian2019benchmark,he2023reinforcement}. In this landscape, the zero-shot learning \cite{dong2023survey} of these models presents a unique advantage, particularly for tasks like meme intervention, which are characterized by data scarcity. \par
While these models hold considerable promise for content moderation, they are not without their limitations. First and foremost, vanilla LLMs and VLMs lack grounding in a knowledge base specific to the task at hand. This can lead to the generation of generic interventions that may fail to adequately address the bias, stereotype, assertions, and toxicity present in the meme content. In terms of visual content, while VLMs have performed well on a variety of traditional visual-linguistic tasks, they often struggle when it comes to memes. The primary reason behind this is the unique nature of memes, which are highly contextual and often rely on a shared understanding of internet subcultures for their interpretation. This makes the accurate assimilation and analysis of memes a challenging task. Finally, even when these models are grounded in a knowledge base for the task of meme intervention, there is a need to filter irrelevant and noisy information. Without appropriate filtering, these models might end up incorporating irrelevant or misleading information into their interventions.\par
In order to address these limitations inherent in LLMs and VLMs for meme intervention, we developed a comprehensive framework called \textit{\textbf{MemeGuard}}. The development of \textit{MemeGuard} was a multi-stage process designed to create a tool capable of understanding and effectively intervening in the spread of toxic memes. In the first stage, we developed a meme-aligned VLM (\textit{\textbf{VLMeme}}), specifically fine-tuned to understand and interpret memes in all their complexity. This allowed our model to delve deeper into the content of memes. Next, we utilized this meme-aligned VLM to identify various facets of the meme, such as the underlying toxicity, bias, stereotypes, and claims being made, which provides valuable insights into the meme's potential harm. To address the challenge of irrelevant knowledge, we then proposed a Multimodal Knowledge Selection mechanism (\textit{\textbf{MKS}}). This mechanism retained only the most relevant knowledge for the intervention generation process. In the subsequent stage, we utilized a general-purpose LLM grounded on this refined knowledge to generate appropriate interventions. This model took the insights provided by the VLM and the ranked knowledge to create contextually relevant and effective responses to toxic memes. Finally, to test the efficacy of our framework, we developed a high-quality labeled dataset featuring a variety of toxic and cyberbully memes with their corresponding human annotated intervention, \textit{\textbf{I}ntervening} \textit{\textbf{C}yberbullying in \textbf{M}ultimodal \textbf{M}emes (ICMM)} dataset. To summarize, we make the following main contributions:\\
\textbf{A novel task} of Meme Intervention to combat the toxicity of cyberbullying memes.\\
\textbf{A novel  dataset}, \textit{ICMM}, to advance the research in this area.\\
\textbf{A novel framework}, \textit{MemeGuard}, that utilizes a meme-aligned VLM (\textit{VLMeme}) to generate contextual information about the meme that is then used to generate the final intervention.
\begin{figure*}[hbt]
	\centering

	\includegraphics[height = 7 cm, origin=c, width = 16cm]{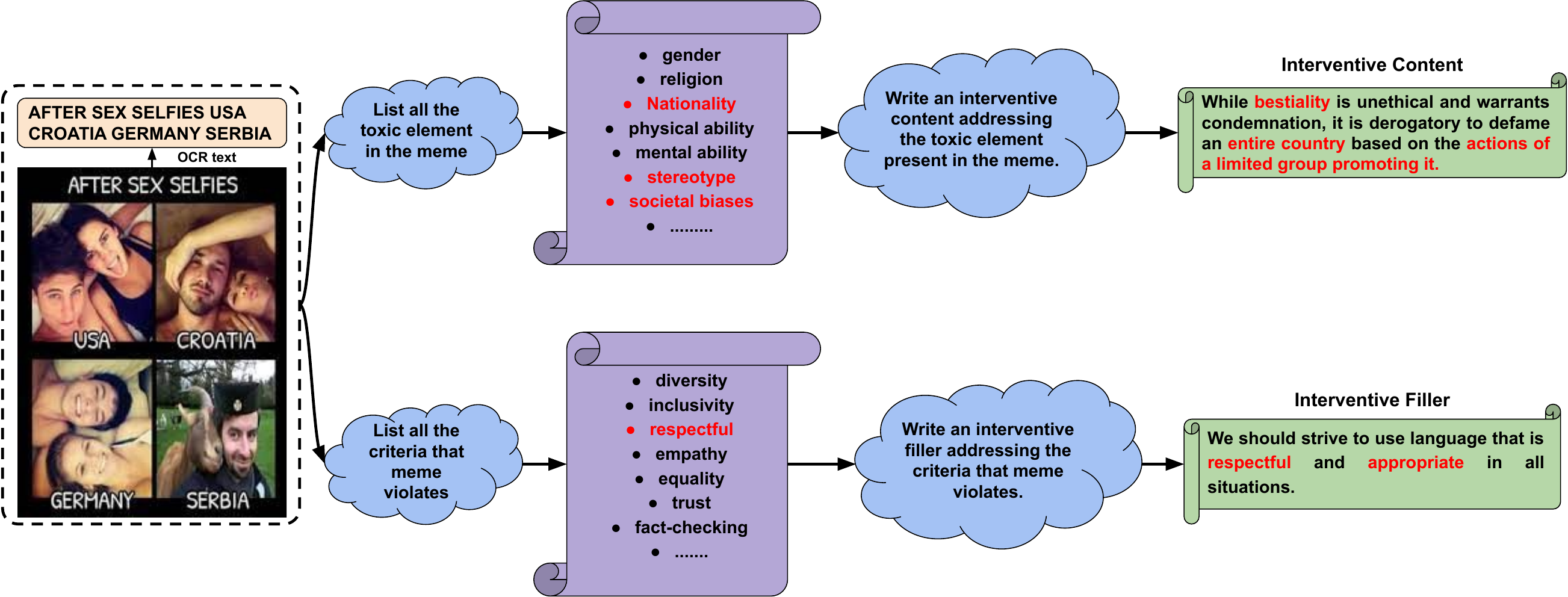}
	\caption {Flowchart depicting the annotation guideline illustrated with a sample example.}
	\label{fig:guide}
 \vspace{-2mm}
\end{figure*} 
\section{Related Works}
\begin{figure*}[hbt]
	\centering
	\includegraphics[height = 10 cm, width = 16cm]{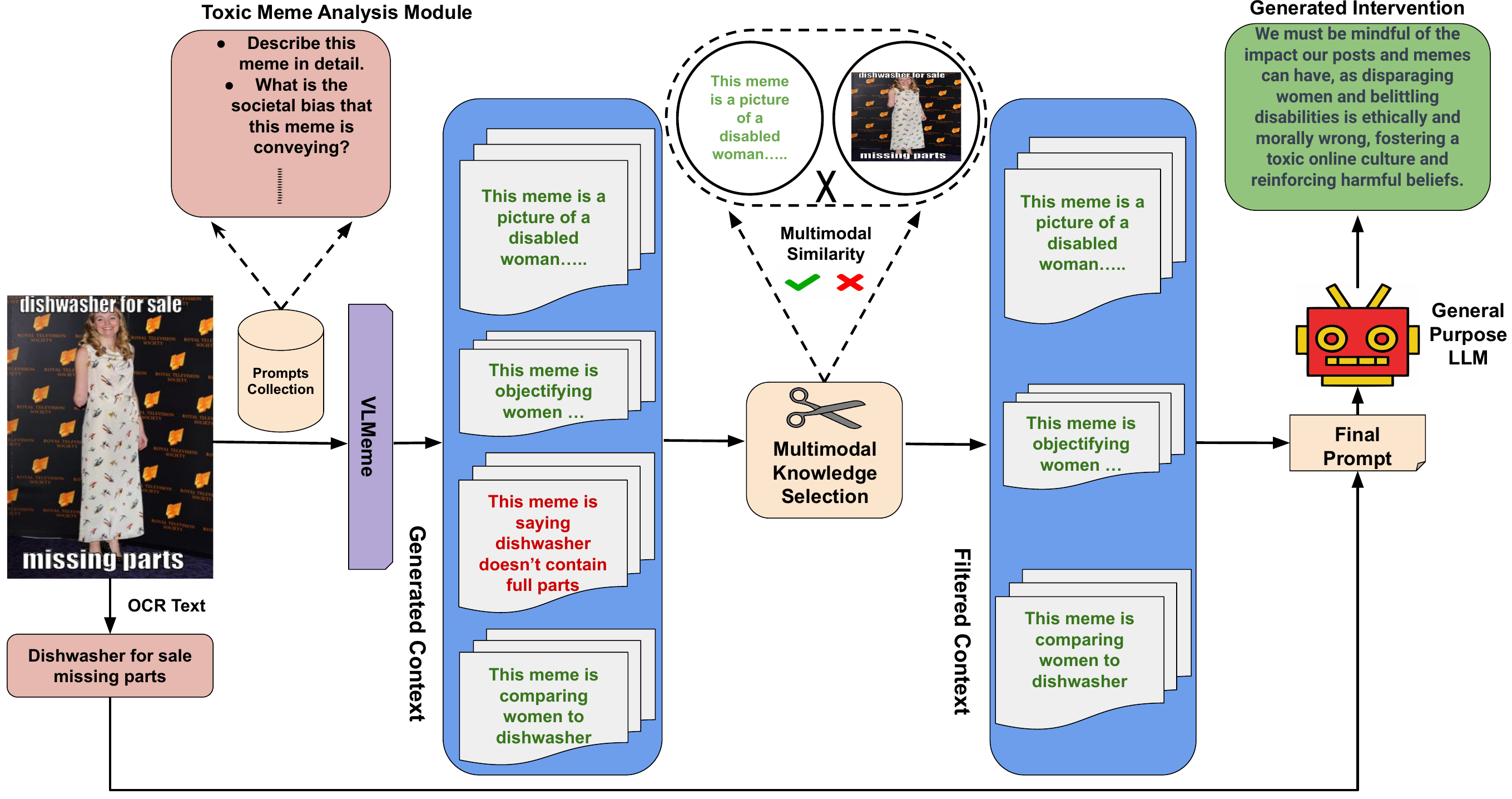}
\vspace{-2mm} 
	\caption {The proposed framework of our \textit{MemeGuard} system. Sentences highlighted in green within the Generated Context block symbolize relevant knowledge, while those in red signify irrelevant knowledge.}
\vspace{-2.5mm} 
	\label{memeG}
\end{figure*} 
\textbf{Meme Analysis: }Meme analysis, a fast-developing field, computationally scrutinizes memes, multimodal entities blending text and visuals, to detect harmful elements like hate speech, offensiveness, cyberbullying, and stereotypes. \citet{kiela2020hateful} proposed a benchmark dataset on hateful memes. ~\citet{pramanick2021momenta} extended the HarMeme dataset~\cite{pramanick2021detecting} with additional memes related to COVID-19 and US politics. ~\citet{10.1145/3477495.3531925} created a cyberbullying meme dataset, which is the only publicly available meme dataset in code-mixed language. Furthermore, ~\citet{mathias2021findings} extended Facebook's meme dataset~\cite{kiela2020hateful} to include two subtasks to identify the attacked category and determine the attack type in memes. Lately, there has been an increase in research dedicated to understanding the context of memes. \citet{hee2023decoding} addressed the research gap by proposing {\em HatReD} dataset annotated with the underlying hateful contextual reasons. Moreover,~\citet{sharma2023memex} proposed a new task that aims to identify evidence from a given context to explain the meme.  \citet{DBLP:conf/eacl/JhaMJVSB24} proposed MultiBully-Ex a benchmark dataset for with multimodal rationales for code-mixed cyberbullying memes. Recently, \citet{hwang2023memecap} also released a labeled dataset for the task of meme caption.\\
\textbf{Intervention and Counterspeech Generation: }
Counterspeech can be regarded as a preferred remedy for combating hate speech as it simultaneously educates the perpetrators about the consequences of their actions and upholds the principles of free speech. Recent studies have shown the remarkable effectiveness of social intervention on social media like Twitter ~\cite{wright2017vectors}, and
Facebook~\cite{schieb2016governing}.  ~\citet{wright2017vectors} studied the conversations on Twitter and found that some arguments between strangers
lead to favorable changes in discourse. ~\citet{mathew2018analyzing} released the first counterspeech dataset with 13,924 manually annotated YouTube comments, marked as counterspeech or not. Due to the limited scalability and generalizability of socially intervened counterspeech datasets, expert-annotated datasets were developed~\cite{qian2019benchmark, chung2019conan} for counterspeech in hate speech.
\section{ICMM Dataset}
To create {\em ICMM}, we utilize {\em MultiBully } dataset~\cite{10.1145/3477495.3531925}, which includes 3222 bully and 2632 nonbully memes. We selected this dataset because it is the only openly available meme dataset on cyberbullying in a code-mixed setting. Our annotation process focuses solely on bully memes to develop interventions against online cyberbullying.
\subsection{Annotation Training}
The annotation team consisted of two expert annotators working in content moderation and three novice annotators, all computer science undergraduate students who possess proficiency in both Hindi and English. First, ten undergraduate computer science students were voluntarily hired through the department email list and compensated through honorarium. 
\textls[-10]{We provided novice annotators with 20 diverse expert-annotated interventions, along with a comprehensive set of annotation guidelines detailed in Appendix~\ref{sec:guide} and illustrated in Figure~\ref{fig:guide} for reference. We have conducted four-phase training to ensure annotators were proficient and well-versed in the tasks. In each training phase, annotators are asked to write interventions for 20 cyberbullying memes. Later, expert annotators assessed the quality of interventions annotated by novice annotators based on fluency, adequacy, informativeness, and persuasiveness as mentioned in Appendix~\ref{sec:ann-quality}. After the completion of each phase of training, expert annotators met with novice annotators to discuss how poorly rated intervention annotations could be improved. These discussions further trained annotators, and simultaneously, annotation guidelines were also renewed. As a result, the top three annotators were selected based on their performance across all phases, whose quality of intervention annotations significantly improved from the first phase (fluency = 3.97, adequacy = 2.59, informativeness = 2.44, and persuasiveness = 2.21) to the fourth phase (fluency = 4.91, adequacy = 4.81, informativeness = 4.79, and persuasiveness = 4.84).}
\subsection{Main Annotation}
We used the open-source platform Docanno\footnote{\url{https://github.com/doccano/doccano}} deployed on a Heroku instance for main annotation (cf. Appendix~\ref{plat}) where three qualified annotators were provided with secure accounts to annotate and track their progress exclusively. Each intervention annotation was assessed by two peer evaluators. The process of annotating interventions is both time-consuming and costly. On average, it takes approximately 6-8 minutes to write interventions for a single meme sample, ensuring high-quality annotation and capturing sufficient evidence and stereotypes for both interventive content and interventive filler and an additional 2-3 minutes for assessment. We offer an honorarium of 8 INR (Indian Rupee) per intervention sample annotation and 2 INR per evaluation, in line with India's minimum wage standards as outlined in the Minimum Wages Act, 1948~\cite{mwa}. As a result of these factors, we have decided to restrict the annotation to only 1000 test samples for this specific project, which is deemed sufficient for testing models based on existing literature. We have commenced annotations for five days a week, adhering to the schedule described in appendix~\ref{sec:schedule}. It took approximately five weeks to complete the entire intervention annotation process. The interventive content has an average length of 14.71 while the interventive filler has an average length of 16.29. In the annotated interventions, we observed remarkably high average scores of 4.98, 4.87, 4.46, and 4.91 for fluency, adequacy, persuasiveness, and informativeness (cf. Appendix~\ref{fig:qual_appen}), respectively. Additionally, two evaluators achieved unanimous agreement scores of 94.7\% for fluency ratings, i.e., evaluators rated the same score for 94.7\% of time, 89.1\% for adequacy ratings, 90.48\% for informativeness ratings, and 82.39\% for persuasiveness ratings.

\section{Methodology}
\textbf{Problem Formulation: }\textls[-10]{The input consists of two entries: (1) the source content $M$, which is a meme potentially bearing toxic content; this multimedia data is composed of visual elements inherent to the image; (2) the OCR text $X$ derived from the source content $M$, representing the textual elements embedded within the meme. The output is a natural language sequence $Y$, which represents the intervention crafted to counter the toxic content of the meme $M$.}\\
This section presents the novel architecture of \textit{\textbf{MemeGuard}} (Figure \ref{memeG}), a multimodal, knowledge-grounded framework designed to mitigate toxic content in memes. For a comprehensive understanding, we partition our approach into three distinct modules: (i) \textbf{Toxic Meme Analysis Module} (ii) \textbf{Multimodal Knowledge Selection (MKS)}, and (iii) \textbf{Intervention Generation Module}.
\subsection{Toxic Meme Analysis Module}

\begin{figure}[hbt]
	\centering
\vspace{-1mm} 
	\includegraphics[height = 4.5cm, width = 5.5cm]{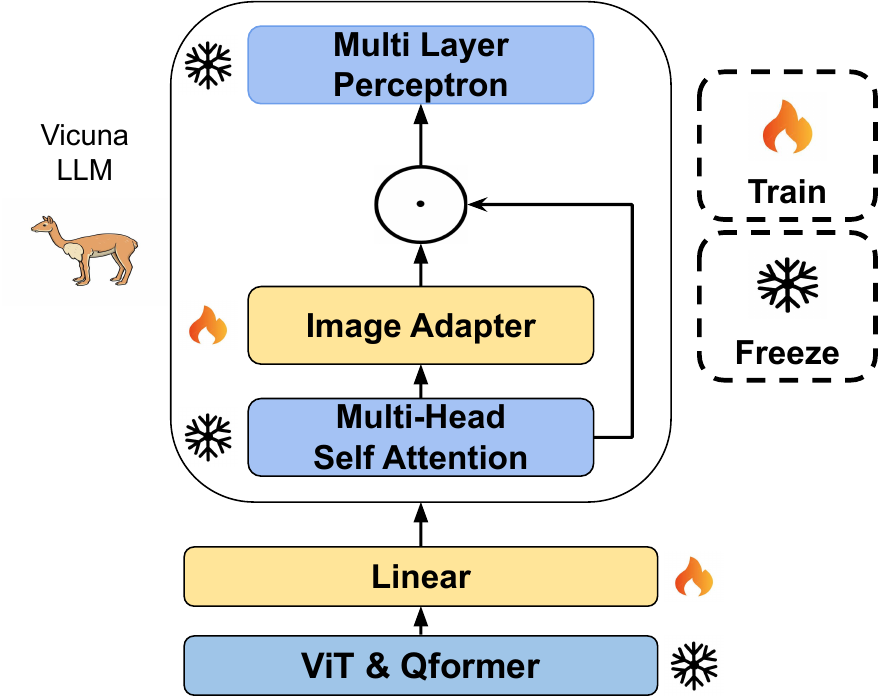}
 \vspace{-1mm}
\caption{Architectural Diagram of \textit{VLMeme}.}
\vspace{-1.5mm}
	\label{VLM}
\end{figure}

\textbf{Development of VLMeme:}  Our model, \textit{\textbf{VLMeme}} (Figure \ref{VLM}), is an enhancement built atop the MiniGPT-4 model \cite{zhu2023minigpt}. MiniGPT-4 is designed to merge visual data from a proficiently trained vision encoder with the capabilities of an advanced large language model (LLM), thereby facilitating complex linguistic tasks. This model leverages the Vicuna language decoder \cite{vicuna2023} -- a model built atop LLaMA \cite{touvron2023llama} -- in tandem with a visual encoder that employs a Vision Transformer (ViT) backbone \cite{dosovitskiy2021image} and a pre-trained Q-Former \cite{zhang2023vision}. 

However, despite their effectiveness in handling traditional visual-linguistic tasks, Visual Language Models (VLMs) tend to face difficulties with meme content due to their inherent complexity and cultural connotations. To surmount this challenge, we augment the Vicuna language decoder with an image adapter inspired by recent works on efficient training of NLP models \cite{houlsby2019parameterefficient,yuan2023artgpt4}, thereby improving its capacity to decipher visual information. Further, we fine-tune this enhanced model on a meme caption dataset \cite{hwang2023memecap}. This approach imbues our model with a deeper understanding of memes, facilitating its handling of the specific nuances and layered meanings that memes often embody. Formally, image adapter computation inside the Vicuna model is shown below:

\vspace{-7mm}
\begin{equation}
\begin{aligned}
\resizebox{0.85\hsize}{!}{%
    $Z_{IA}=Linear_{1}(RELU(Linear_2(Z_{MHA})))+Z_{MHA}$
}
\end{aligned}
\end{equation}
\vspace{-6mm}

where $Z_{MHA}$ represents the output from Multi-Headed Attention layer of Vicuna, $Z_{IA}$ represents the output from this image adapter layer,  RELU represents the activation function computations, $Linear_{1}$, and $Linear_{2}$ represent the linear feed-forward neural network layers. Following the incorporation of the image adapter within the Vicuna language decoder, the subsequent step involves fine-tuning this enhanced model on a meme dataset. This dataset is composed of memes accompanied by their respective descriptions, serving as a rich resource for context-based meme understanding. For this purpose, we exploit the recently released \textit{MEMECAP} \cite{hwang2023memecap} dataset, which offers a wide range of memes with their associated descriptions. The fine-tuning strategy aligns with the method described in \citet{zhu2023minigpt,ghosh2024medsumm}. During this process, the components of Vicuna, Q-Former, and ViT are kept frozen, allowing us to focus specifically on the meme understanding abilities of our model. This will result in a meme-aligned vision language model \textit{\textbf{VLMeme}}.\\
\textbf{Contextual Knowledge Generation:} The next stage in the development of our framework involves utilizing \textit{\textbf{VLMeme}} to generate contextual knowledge to gain a deeper understanding of the meme content, allowing it to identify the underlying toxicity, biases, stereotypes, and assertions present within the meme. Formally, for a given meme $M$, we formulate a set of prompts $P=\{p_0, p_1, \ldots, p_n\}$ with the aim of generating diverse contextual information about meme $M$, $KS=\{ks_0, ks_1, \ldots, ks_n\}$ by prompting \textit{\textbf{VLMeme}} as follows: for each prompt $p_i$: $ks_i=\textit{VLMeme}(p_i)$. Specifically, these prompts are designed to yield the following critical insights: (i) meme description, (ii) bias inherent in the meme, (iii) stereotypes propagated by the meme, (iv) toxic elements within the meme, and (v) assertions and claims conveyed through the meme. The detailed set of prompts is presented below.
\begin{tcolorbox}[left=10pt,right=2pt,size=small,colback=blue!5!white,colframe=blue!75!black,title=Prompts used to generate contextual information,fonttitle=\bfseries\fontsize{9}{9.6}\selectfont]
  \begin{itemize}
{\fontfamily{phv}\fontsize{7.5}{8.5}\selectfont
    \setlength\itemsep{0em}  
      \item Describe this meme in detail.
      \item What is the societal bias that this meme is conveying?
      \item What is the societal stereotype that this meme is conveying?
      \item What is the toxicity and hate that this meme is spreading?
      \vspace{-2mm}
      \item What are the claims that this meme is making?
}    
  \end{itemize}
\end{tcolorbox}

\begin{table*}[]
\centering
\scalebox{0.60}{%
\begin{tabular}{c|c|ccc|cccc|cc|c|}
\cline{2-12}
 &
  \multirow{2}{*}{\textbf{Model}} &
  \multicolumn{3}{c|}{\textbf{ROUGE}} &
  \multicolumn{4}{c|}{\textbf{BLEU}} &
  \multicolumn{2}{c|}{\textbf{Average Score}} &
  \multirow{2}{*}{\textbf{BERTScore}} \\ \cline{3-11}
                                            &                & \textbf{R1}    & \textbf{R2}   & \textbf{RL}    & \textbf{B1}    & \textbf{B2}    & \textbf{B3}    & \textbf{B4}    & \textbf{BLEUavg} & \textbf{Hmean}        &       \\ \hline
\multicolumn{1}{|c|}{\multirow{4}{*}{{\textbf{Dolly}}}} &
  OCR &
  2.39 &
  0.15 &
  2.07 &
  7.49 &
  4.52 &
  1.84 &
  0.84 &
  3.67 &
  2.65 &
  74.35 \\
\multicolumn{1}{|c|}{}                      & OCR + MiniGPT4 & 0.94  & 0.05 & 0.74  & 2.77  & 1.73  & 0.75  & 0.36  & 1.40 & 0.97 & 73.81     \\
\multicolumn{1}{|c|}{}                      & OCR + VLmeme   & 8.48  & 0.23 & 6.26  & 16.55 & 12.11 & 6.17  & \textbf{3.36}  & 9.55  & 7.56  & 79.17 \\
\multicolumn{1}{|c|}{}                      & MemeGuard      & \textbf{8.68}  & \textbf{0.41} & \textbf{7.22}  & \textbf{17.05} & \textbf{12.88} & \textbf{6.65}  & 2.86  & \textbf{9.86}    & \textbf{8.33}  & \textbf{82.27} \\ \hline 
\multicolumn{1}{|c|}{\multirow{4}{*}{{\textbf{LLaMA}}}} &
  OCR &
  8.48 &
  0.85 &
  6.96 &
  31.15 &
  27.7 &
  10.56 &
  4.6 &
  18.50 &
  \textbf{10.11} &
  79.98 \\
\multicolumn{1}{|c|}{}                      & OCR + MiniGPT4 & 8.84  & 0.87 & \textbf{6.98}  & 31.19 & 27.77 & 10.64 & 4.74  & 18.58 & 10.15  & 79.98 \\
\multicolumn{1}{|c|}{}                      & OCR + VLmeme   & 6.18  & 0.61 & 4.73  & 30.89 & 28.4  & 10.95 & \textbf{4.78}  & 18.75  & 7.55   & 80.11 \\
\multicolumn{1}{|c|}{}                      & MemeGuard      & \textbf{9.27}  & \textbf{1.71} & 5.29  & \textbf{31.96} & \textbf{29.08} & \textbf{11.89} & 4.3   &\textbf{19.31} & 8.30  & \textbf{80.11} \\ \hline 
\multicolumn{1}{|c|}{\multirow{4}{*}{{\textbf{RedPajama}}}} &
  OCR &
  8.31 &
  0.45 &
  6.63 &
  \textbf{39.17} &
  19.9 &
  7.47 &
  3.34 &
  17.47 &
  9.61 &
  79.56 \\
\multicolumn{1}{|c|}{}                      & OCR + MiniGPT4 & \textbf{11.61} & 0.7  & 8.55  & 33.94 & 22.19 & 10.6  & 5.2   & 17.98 & 11.58  & 82.4  \\
\multicolumn{1}{|c|}{}                      & OCR + VLmeme   & 10.82 & \textbf{0.58} & 7.93  & 35.1  & 22.72 & 10.02 & 4.88  & 18.18   & 11.04  & 80.93 \\
\multicolumn{1}{|c|}{}                      & MemeGuard      & 10.83 & 0.57 & \textbf{9.01}  & 35.23 & \textbf{22.82} & \textbf{11.05} & \textbf{4.9}   & \textbf{18.5}    & \textbf{12.11}  & \textbf{83.42} \\ \hline 
\multicolumn{1}{|c|}{\multirow{4}{*}{{\textbf{FLAN-T5}}}} & OCR            & 5.92  & 0.4  & 4.96  & 10.78 & 4.16  & 1.34  & 0.59  & 4.22  & 4.55 & 83.19   \\
\multicolumn{1}{|c|}{}                      & OCR + MiniGPT4 & 13.88 & 1.03 & 10.13 & 39.65 & 26.68 & 12.75 & 6.58  & 21.41  & 13.75  & 84.5  \\
\multicolumn{1}{|c|}{}                      & OCR + VLmeme   & 13.96 & 0.77 & 10.91 & 45.47 & 27.98 & 12.33 & \textbf{7.22}  & 23.25   & 14.85 & 85.04 \\
\multicolumn{1}{|c|}{}                      & MemeGuard      & \textbf{14.11} & \textbf{2.01} & \textbf{11.22} & \textbf{48.91} & \textbf{30.21} & \textbf{12.81} & 6.6   & \textbf{24.63} & \textbf{15.41}  & \textbf{87.21} \\ \hline 
\multicolumn{1}{|c|}{\multirow{4}{*}{{\textbf{GPT3.5-Turbo}}}} &
  OCR &
  21.05 &
  3.85 &
  13.25 &
  37.9 &
  29.57 &
  17.53 &
  11.24 &
  24.06 &
  17.08 &
  85.07 \\
\multicolumn{1}{|c|}{}                      & OCR + MiniGPT4 & 13.71 & 1.83 & 9.3   & 21.3  & 17.99 & 10.97 & 6.51  & 14.19 & 11.23  & 85.33 \\
\multicolumn{1}{|c|}{}                      & OCR + VLmeme   & 22.44 & 3.13 & 10.18 & 37.76 & 30.23 & 17.53 & 11.35 & 24.22 & 14.33  & 86.26 \\
\multicolumn{1}{|c|}{}                      & MemeGuard      & \textbf{24.42} & \textbf{4.7}  & \textbf{15.22} & \textbf{39.45} &\textbf{32.32} & \textbf{17.82} & \textbf{11.91} & \textbf{25.37}  & \textbf{19.03}  & \textbf{89.02} \\ \hline
\end{tabular}%
   }
\vspace{-1.5mm}
\caption{Performances of various \textit{MemeGuard} models and corresponding baselines, evaluated using automatic metrics with different base LLMs.}
\label{tab:quant}
\vspace{-2.5mm}
\end{table*}
\vspace{-1.5mm}

\subsection{Multimodal Knowledge Selection (MKS)}

\vspace{-1mm}

\begin{table}[]
\centering
\scalebox{0.50}{%
\begin{tabular}{c|c|c|c|c|c|}
\cline{2-6}
                                                    & \textbf{Model}          & \textbf{Fluency} & \textbf{Adequacy} & \textbf{Persuasiveness} & \textbf{Informativeness} \\ \hline
\multicolumn{1}{|c|}{\multirow{4}{*}{\textbf{FLAN-T5}}}         & OCR            & 4.27    & 2.69     & 2.16           & 2.72            \\
\multicolumn{1}{|c|}{}                              & OCR + MiniGPT4 &   4.31      &    2.81      &       2.19        &       2.79          \\
\multicolumn{1}{|c|}{}                              & OCR + VLMeme   &   4.35      &       3.26   &     2.23           &     3.19            \\
\multicolumn{1}{|c|}{}                              & MemeGuard      & 4.39    & 3.32     & 2.24           & 3.27            \\ \hline
\multicolumn{1}{|c|}{\multirow{4}{*}{\textbf{GPT3.5-Turbo}}} & OCR            & 4.79    & 3.6      & 3.32           & 3.57            \\
\multicolumn{1}{|c|}{}                              & OCR + MiniGPT4 &     4.81    &   3.82       &       3.47         &     3.61            \\
\multicolumn{1}{|c|}{}                              & OCR + VLMeme   &  4.81       &      3.95    &        3.91        &     4.05           \\
\multicolumn{1}{|c|}{}                              & MemeGuard      & 4.82    & 4.16     & 3.97           & 4.13            \\ \hline
\multicolumn{2}{|c|}{Annotated Intervention} & 4.98 & 4.87 & 4.46 & 4.91 \\ \hline
\end{tabular}%
}
\vspace{-1.5mm}
\caption{\textls[0]{Human evaluation scores of the two best \textit{MemeGuard} models and their corresponding baselines across different metrics.}}
\label{tab:human}
\vspace{-4mm}
\end{table}

\textls[-10]{While we anticipate that \textit{\textbf{VLMeme}} will generate relevant information, $KS$, there are instances where these models might deviate from the main query \cite{holtzman2020curious}. This could introduce irrelevant context, distracting the Large Language Models (LLMs) \cite{shi2023large}, thereby hampering its ability to accurately analyze the meme's content and generate effective interventions. As such, there is a need to filter out this irrelevant context to maintain the effectiveness of the intervention process. To address this issue, we propose a filtering strategy named \textit{Multimodal Knowledge Selection}, which works as follows.
Formally, for each $ks_i$, which corresponds to a specific text field in the meme, it is further broken down into a set of $m$ sentences $ks_i=\{s_1, s_2, \ldots, s_m\}$ using sentence tokenization\footnote{\url{https://www.nltk.org/api/nltk.tokenize.html}}. 
To achieve this, we employ a pre-trained multimodal model, ImageBind \cite{girdhar2023imagebind}, which serves as an off-the-shelf encoder-based multimodal model $enc(\cdot)$ that maps a token sequence (text) and an image to their respective feature vectors embedded in a unified vector space. Cosine similarity $sim(\cdot, \cdot)$ is then used to measure the relevance of each sentence to the image. Formally, a sentence $s_j$ is retained if $sim(enc(s_j), enc(M)) > Th$, where $M$ is the image associated with the meme, and $Th$ is the predefined similarity threshold. Building upon this, we define the subset of retained knowledge sentences, having undergone the \textit{MKS}, as follows:}

\vspace{-6mm}
\begin{equation}
\begin{aligned}
\resizebox{0.89\hsize}{!}{$
    ks'_i = \{s_j | s_j \in ks_i \,\text{and } sim(enc(s_j), enc(M)) > Th\}$
}
\end{aligned}
\end{equation}
\vspace{-5.5mm}

\textls[-10]{With this approach, the filtered knowledge sentences $KS'=\{ks'_0, \ldots, ks'_n\}$ not only hold a high degree of visual-textual alignment with the corresponding meme but are also contextually relevant.
\subsection{Intervention Generation Module}
This module is designed to utilize the refined knowledge sentences $KS'=\{ks'_0, ks'_1, \ldots, ks'_n\}$ to generate well-informed and contextually pertinent interventions for toxic memes. The Intervention Generation Module operates on a pre-trained Large Language Model (LLM), denoted as $LM$. To generate the intervention, the $LM$ is prompted with a specially designed prompt $P_{in}$, which incorporates both the meme's OCR text $X$ and the generated knowledge about the meme $KS'$. The output of this process, $I$, is the intervention text generated by the model. Formally, this process can be described as follows: $I = LM(P_{in}(X, KS'))$.
The detailed prompt is presented below.}

\vspace{-1mm}
\begin{tcolorbox}[size=small,colback=blue!5!white,colframe=blue!75!black,title=Prompt used to generate final intervention,fonttitle=\bfseries\fontsize{9}{9.6}\selectfont]
{\fontfamily{phv}\fontsize{7.5}{8.75}\selectfont
    \setlength\itemsep{0em}  
    This is a toxic meme with the description: $\{ks^{'}_{0}\}$. The following text is written inside the meme: $\{X\}$.  Rationale: Bias: $\{ks^{'}_{1}\}$, Toxicity: $\{ks^{'}_{2}\}$, Claims: $\{ks^{'}_{3}\}$, and Stereotypes: $\{ks^{'}_{4}\}$.  Write an intervention for this meme based on all this knowledge.
}    
\end{tcolorbox}
\vspace{-2mm}

\section{Experimental Results and Discussion}
\vspace{-1mm}
\textbf{Experimental Setup: }We leveraged following general purpose LLMs for intervention generation module: Dolly\footnote{https://huggingface.co/databricks/dolly-v2-3b}, LLaMA~\cite{touvron2023llama}, RedPajama\footnote{https://huggingface.co/togethercomputer/RedPajama-INCITE-Chat-3B-v1}, FLAN-T5~\cite{chung2022scaling}, and GPT3.5-Turbo. Details are mentioned in Appendix Table \ref{LLMs}. We tested these LLMs on different settings: (1) With only OCR text in prompt (OCR), With the knowledge obtained from MiniGPT4 (OCR+MiniGPT4), With the knowledge obtained from VLMeme (OCR+VLMeme), and then our proposed framework (\textit{MemeGuard}). We have set the threshold parameter $T$ to 0.5 and temperature to 0.5 across all settings (cf. Appendix \ref{hype} for details). We utilize ROUGE ~\cite{lin2004rouge}, BLEU~\cite{papineni2002bleu},  BLEUavg: the average of BLEU score, Hmean: harmonic mean of ROUGE-L and BLEU, and BERTScore~\cite{bert-score} as automatic evaluation metrics. For the human evaluation, we enlist three qualified human evaluators who assess the generated interventions based on four aspects: Fluency: Grammatical Correctness, Adequacy: Relevance to the meme, Persuasiveness: fluency, adequacy, persuasiveness, and informativeness (cf. Appendix~\ref{sec:eval}).\\

\begin{table*}[]
\scalebox{0.435}{
\begin{tabular}{|cc|c|c|}
\hline
\multicolumn{2}{|c|}{Hateful Memes}                                       & (A) \begin{minipage}{0.78\textwidth}
\centering
      \includegraphics[width=\linewidth, height=60mm]{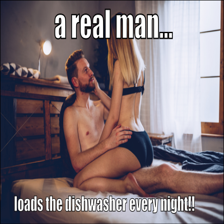}
    \end{minipage}                                & (B) \begin{minipage}{0.75\textwidth}
\centering
      \includegraphics[width=\linewidth, height=60mm]{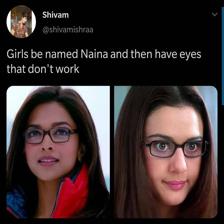}
    \end{minipage}       \\ \hline
\multicolumn{2}{|c|}{OCR Text}                                            &\begin{tabular}[c]{@{}l@{}} a real man... loads the dishwasher every night!! \end{tabular}  & \begin{tabular}[c]{@{}l@{}} Girls be named Naina and then have eyes that don't work \end{tabular}  \\ \hline
\multicolumn{1}{|c|}{\multirow{4}{*}{FLAN T5}}      & OCR Text            & \begin{tabular}[c]{@{}l@{}}Consistency in loading the dishwasher daily maintains cleanliness, but it's\\ equally vital to consider individual circumstances and needs in setting routines.\end{tabular} & \begin{tabular}[c]{@{}l@{}} Naina, you are a girl with eyes that don't work. \end{tabular} \\ \cline{2-4} 
\multicolumn{1}{|c|}{}                              & OCR Text + MiniGPT4 & \begin{tabular}[c]{@{}l@{}}The meme is not making any claims, as it is simply depicting a light-hearted and playful moment\\ between a man and a woman. The meme is not making any political or social statements, nor is it\\ making any scientific or historical claims. It is simply a fun and entertaining image that is designed to\\ bring a smile to people's faces and brighten up their day. Therefore, the meme is not making any claims.\end{tabular}  &  \begin{tabular}[c]{@{}l@{}} I'm sorry, but I don't have the information you provided earlier with the meme.\\ Can you please provide me with the meme again so that I can describe it in detail? \end{tabular}\\ \cline{2-4} 
\multicolumn{1}{|c|}{}                              & OCR Text + VLMeme   &\begin{tabular}[c]{@{}l@{}} This meme is spreading the toxicity and hate that men are not interested in doing household chores\\ and that they think it's the woman's responsibility, and this is perpetuating gender stereotypes.\end{tabular}  &  \begin{tabular}[c]{@{}l@{}} I apologize but I do not see any toxicity or hate in the meme. The meme is a simple\\ observation about the similarity between two people named Nina, and how they\\ both wear glasses but don't know their names. It does not promote any kind of\\ harmful behavior or discrimination. \end{tabular}\\ \cline{2-4} 
\multicolumn{1}{|c|}{}                              & MemeGuard           &  \begin{tabular}[c]{@{}l@{}} 
This meme reinforces harmful stereotypes about men and their roles in the household. A real man \\is not defined by their ability or willingness to load the dishwasher; it is a household chore equally\\ shared between partners. Interventions could include addressing the underlying stereotypes and\\ promoting shared responsibility and respect in domestic relationships.\end{tabular}  & \begin{tabular}[c]{@{}l@{}} I apologize but I do not see any toxicity or hate in the meme.\end{tabular} \\ \hline
\multicolumn{1}{|c|}{\multirow{4}{*}{GPT3.5-Turbo}} & OCR Text            &\begin{tabular}[c]{@{}l@{}} It is important to recognize that tasks such as loading the dishwasher should not have gender roles\\ attached to them. Both men and women should be responsible for household chores, and it should not\\ be labeled as a "real man" task. Rather, all individuals living together in a household could take turns\\ in completing these tasks, creating a more equal and fair division of labor. Gender stereotypes limit our\\ potential and should be challenged.\end{tabular}  & \begin{tabular}[c]{@{}l@{}} Let's break stereotypes and celebrate individuality. Naming someone 'Naina'\\ doesn't determine their vision. Embrace diversity and remember names do not\\ define abilities. \#Empowerment \#NoMoreLabels
\end{tabular}  \\ \cline{2-4} 
\multicolumn{1}{|c|}{}                              & OCR Text + MiniGPT4 &\begin{tabular}[c]{@{}l@{}} We would like to remind everyone that posting toxic memes that perpetuate harmful gender stereotypes \\is not only hurtful but also damaging to our society. The recent meme suggesting that men don't care\\ about doing household chores is not only untrue but also perpetuates harmful gender stereotypes.\\ We urge you to consider the impact of your words and actions before posting such content.  \end{tabular}   &  \begin{tabular}[c]{@{}l@{}} Hey there, memes should be created with the intention of spreading positivity\\ and inclusivity. Let's focus on sharing content that brings people together\\ rather than perpetuating stereotypes or potentially hurtful jokes.
\end{tabular} \\ \cline{2-4} 
\multicolumn{1}{|c|}{}                              & OCR Text + VLMeme   &\begin{tabular}[c]{@{}l@{}} It's important to recognize that household responsibilities should be shared between partners regardless \\of gender. Posting memes that perpetuate stereotypes and promote toxic gender roles only contribute\\ to division and inequality within relationships. Let's encourage inclusive and equal participation in \\household chores for healthier and happier partnerships.\end{tabular}  & \begin{tabular}[c]{@{}l@{}} It's important to remember that people's identities go beyond their physical \\appearance or accessories. Posting memes that perpetuate stereotypes and\\ focus on superficial traits undermines the value of personal identity and can\\ contribute to harmful biases. Let's strive to create content that celebrates\\ diversity and promotes respect for individuals' unique qualities.
\end{tabular}  \\ \cline{2-4} 
\multicolumn{1}{|c|}{}                              & MemeGuard           &\begin{tabular}[c]{@{}l@{}}
  Posting memes that reinforce gender stereotypes and biases can be harmful and contribute to a toxic\\ culture. Let's challenge stereotypes and biases, and encourage others to do the same. The next time \\you come across a meme like this, take a moment to reflect on its message and consider the harm\\ it may cause. Let's use social media as a tool to spread positivity and inclusivity, and promote a \\more accepting and equitable world.
  
  \end{tabular}  &  \begin{tabular}[c]{@{}l@{}} It is important to recognize that memes like this reinforce harmful\\ stereotypes and perpetuate biases. We should aim to promote inclusivity\\ and respect by creating and sharing content that celebrates personal \\identity and discourages judgment based on appearance or accessories.
\end{tabular} \\ \hline
\multicolumn{2}{|c|}{Annotated Intervention}                              &  \begin{tabular}[c]{@{}l@{}}Defining rigid gender roles that prescribe certain behaviors for "real men" and dehumanizing women\\ by reducing them to mere sexual objects or limiting their worth to domestic chores perpetuates \\harmful stereotypes. Promoting equality in household chores can lead to healthier\\ and more balanced relationships, benefiting both partners.\end{tabular} &\begin{tabular}[c]{@{}l@{}} Making fun of someone's name or physical
ability is disrespectful and hurtful.\\
We should strive to treat others with kindness
and empathy, recognizing the\\ diversity and
beauty of all individuals\end{tabular}  \\ \hline
\end{tabular}
}
\caption{Sample interventions  generated by best two \textit{MemeGuard} models and their corresponding baselines.}
\label{fig:qual_appen}
\end{table*}
\vspace{-1mm}
\textbf{Automatic Evaluation: }\textls[-5]{Table \ref{tab:quant} presents the results of the automatic evaluation on our framework. By integrating MiniGPT4, there are notable enhancements in RedPajamas and FLAN-T5's performance, evidenced by a 1.97 and 9.2 point increase, respectively, in Hmean metric, and significant BERTScore boosts of 2.84 and 84.5 points, respectively. This underlines the benefit of added image context for generating improved interventions. However, MiniGPT4's generated knowledge may not be entirely relevant, causing only slight BERTScore increases in LLaMA and GPT3.5-Turbo, as they can be sensitive to prompt noise \cite{shi2023large}. Dolly's performance didn't improve with the incorporation of MiniGPT4.\\
Substituting MiniGPT4 with VLMeme has led to pronounced enhancements, particularly in the cases of FLAN-T5, GPT3.5-Turbo, and Dolly, as reflected in their Hmean and BERTScore. This suggests that VLMeme's ability to generate more contextually relevant knowledge about memes than MiniGPT4 may account for this improvement.
The outcome underscores the power of domain-specific fine-tuning in boosting the effectiveness of vision-language models. We delve into a performance analysis of VLMeme for the meme comprehension task in Appendix \ref{vlmsec}.}
When our proposed framework, {\em MemeGuard}, which features the incorporation of MKS over VLMeme, is utilized by Dolly, LLaMA, RedPajamas, FLAN, and GPT3.5-Turbo, it outperforms all the baselines across all automatic evaluation metrics. This illustrates the beneficial effect of maintaining only pertinent and crucial knowledge in the prompt, enhancing in-context learning for these LLMs. Using {\em MemeGuard}, Dolly, LLaMA, RedPajamas, FLAN, and GPT3.5-Turbo achieved Rouge-L scores of 7.22, 5.29, 9.01, 11.22, and 15.22, respectively. The average BLEU scores for these models are 9.86, 19.31, 18.5, 24.63, and 25.37, and the BERTScores are 82.27, 80.11, 83.42, 87.21, and 89.02, respectively.\\
FLAN-T5 and GPT3.5-Turbo, when leveraging \textit{MemeGuard}'s framework, outshine other models in ROUGE, BLEU, and BERTScore metrics. Yet, they only achieve low N-gram matching scores, with Hmean scores of 15.41 and 19.03, respectively. Despite this, their BERTScores remain fairly high (GPT3.5-Turbo at 89.02, FLAN-T5 at 87.21), suggesting generated interventions, while differing in word choice, still convey meanings similar to the ground truth. These findings prompted us to initiate a human evaluation as outlined in Section~\ref{sec:eval}.


\textbf{Human Evaluation:} 
Table~\ref{tab:human} presents the results of the human evaluation conducted on the generated interventions using GPT3.5-Turbo and FLAN-T5, leveraging {\em MemeGuard } and their corresponding baselines. According to the assessments made by the human evaluators, FLAN-T5 exhibits a high level of fluency (4.39) in terms of language proficiency. However, it falls short in terms of adequacy (3.32), persuasiveness (2.24), and informativeness (3.27). On the other hand, GPT3.5-Turbo achieves the highest scores for fluency (4.82), adequacy (4.16), persuasiveness (3.97), and informativeness (4.13) in human evaluation. It can also be seen how the { \em MemeGuard } framework consistently outperforms  OCR, OCR + MiniGPT4, OCR + VLMeme counterparts across all metrics. However, the human-annotated interventions still demonstrate higher quality, with fluency rated at 4.98, adequacy at 4.87, persuasiveness at 4.46, and informativeness at 4.91. These findings underscore the challenges faced by FLAN-T5 and GPT3.5-Turbo in generating interventions that fully meet the desired criteria, especially when compared to the quality of the annotated interventions.\\
\textbf{Qualitative Analysis: }
\textls[-10]{Table~\ref{fig:qual_appen} presents examples of interventions generated by both FLAN-T5 and GPT3.5-Turbo utilizing {\em MemeGuard } and their corresponding baselines. In example (A), FLAN-T5 with OCR Text and OCR Text + MiniGPT4 fail to recognize the hatefulness of the meme. FLAN-T5 with OCR + VLMeme recognizes the gender stereotypes present in the meme but doesn't acknowledge the cause of gender stereotypes. FLAN-T5 with {\em MemeGuard } framework captures the "gender roles" associated with the meme and explicitly acknowledges them using the words "real men" and "willingness to load." On the other hand, GPT3.5-Turbo seems to have excellent language modeling capability, as it can recognize the gender stereotypes present in the meme with reasoning capability enhancing as we go from OCR Text, OCR Text + MiniGPT4, OCR Text + VLMeme, and {\em MemeGuard}. However, both GPT3.5-Turbo and FLAN-T5 do not acknowledge the gender association of the word "dishwasher" with women and the sexual objectification of women present in the meme, as mentioned in the Annotated Intervention. In example (B), we can observe that FLANT5 fails to understand the sarcastic wordplay between  ``Naina" (a Hindi word written in English script) and ``eyes don't work". Hence, it doesn't recognize the hatefulness of the meme. Even after being supplied with knowledge from {\em MemeGuard}, it fails to understand the meme. This indicates a lack of interpretation of code-mixed text. Meanwhile, GPT3.5-Turbo can recognize the meaning of the word "Naina", i.e., "vision". It can also be seen that the reasoning of intervention generation improves as the knowledge provided enhances from only OCR Text to {\em MemeGuard}. However, in all cases, GPT3.5-Turbo highlights the focus on "appearance" or "accessories" rather than "physical ability," which is mocked in the meme as mentioned in Annotated Intervention. }

\vspace{-2mm}
\section{Conclusion and Future Work}
\vspace{-1mm}
In this paper, we introduce ICMM, the first meme dataset incorporating human-annotated interventions for 1000 cyberbullying memes, setting a new gold standard. The ICMM dataset presents novel avenues for assessing the efficacy of interventions generated by generative models relative to human responses. We further put forth the {\em MemeGuard}  framework, harnessing the power of Large Language Models (LLMs) and Visual Language Models (VLMs) in the generation of meme interventions, supplemented by the introduction of a meme-focused Vision Language model (VLMeme) and a Multimodal Knowledge Selection mechanism (MKS). In future research, we aim to focus on the development of more resilient and effective vision-language models for a better understanding of memes. Additionally, we plan to expand this work to encompass toxic videos and reels.

\section*{Limitations}

Despite the considerable accomplishments achieved in our study, we must acknowledge several limitations that warrant attention. Firstly, the scope of prompt selection was restricted predominantly to few-shot prompting strategies, omitting a thorough exploration of the impact that prompt variation could potentially have on performance outcomes. As such, this represents an area for potential future investigation. Secondly, while our VLMeme model demonstrates promising proficiency in meme understanding, it is not without its shortcomings. There are instances where the model fails to accurately interpret the memes, suggesting room for enhancement and adaptation. From Table~\ref{fig:qual_appen} example (B), it is also clear that the opensource LLM, i.e., FLAN-T5, fails to handle code-mixing present inside the text. Hence, apart from the design of {\em MemeGuard},  the overall model's performance also depends on the language modeling ability of the LLMs across languages, as seen in the case of code-mixing. Lastly, the creation of a dataset for meme interpretation posed certain challenges, primarily related to cost. As a result, the dataset used in our study was limited to just 1000 instances. These 1000 instances consist of code-mixed memes in Indian languages as well as plain English memes in a Western context. Therefore, it limits the evaluation of the proposed framework in the cultural context of English and code-mixed Indian memes. Consequently, future work may wish to consider methods of expanding the dataset to provide a more robust basis for model evaluation and refinement across different languages and cultures globally.


\bibliography{custom,anthology}

\begin{thebibliography}{50}
\expandafter\ifx\csname natexlab\endcsname\relax\def\natexlab#1{#1}\fi

\bibitem[{mwa()}]{mwa}

\newblock \url{https://en.wikipedia.org/wiki/List_of_countries_by_minimum_wage}.

\bibitem[{res()}]{rescale}

\newblock \url{https://github.com/Tiiiger/bert_score/blob/master/journal/rescale_baseline.md}.

\bibitem[{Chavan et~al.(2023)Chavan, Liu, Gupta, Xing, and Shen}]{chavan2023one}
Arnav Chavan, Zhuang Liu, Deepak Gupta, Eric Xing, and Zhiqiang Shen. 2023.
\newblock One-for-all: Generalized lora for parameter-efficient fine-tuning.
\newblock \emph{arXiv preprint arXiv:2306.07967}.

\bibitem[{Chiang et~al.(2023)Chiang, Li, Lin, Sheng, Wu, Zhang, Zheng, Zhuang, Zhuang, Gonzalez, Stoica, and Xing}]{vicuna2023}
Wei-Lin Chiang, Zhuohan Li, Zi~Lin, Ying Sheng, Zhanghao Wu, Hao Zhang, Lianmin Zheng, Siyuan Zhuang, Yonghao Zhuang, Joseph~E. Gonzalez, Ion Stoica, and Eric~P. Xing. 2023.
\newblock \href {https://lmsys.org/blog/2023-03-30-vicuna/} {Vicuna: An open-source chatbot impressing gpt-4 with 90\%* chatgpt quality}.

\bibitem[{Chung et~al.(2022)Chung, Hou, Longpre, Zoph, Tay, Fedus, Li, Wang, Dehghani, Brahma et~al.}]{chung2022scaling}
Hyung~Won Chung, Le~Hou, Shayne Longpre, Barret Zoph, Yi~Tay, William Fedus, Eric Li, Xuezhi Wang, Mostafa Dehghani, Siddhartha Brahma, et~al. 2022.
\newblock Scaling instruction-finetuned language models.
\newblock \emph{arXiv preprint arXiv:2210.11416}.

\bibitem[{Chung et~al.(2019)Chung, Kuzmenko, Tekiroglu, and Guerini}]{chung2019conan}
Yi-Ling Chung, Elizaveta Kuzmenko, Serra~Sinem Tekiroglu, and Marco Guerini. 2019.
\newblock Conan--counter narratives through nichesourcing: a multilingual dataset of responses to fight online hate speech.
\newblock \emph{arXiv preprint arXiv:1910.03270}.

\bibitem[{Dettmers et~al.(2023)Dettmers, Pagnoni, Holtzman, and Zettlemoyer}]{dettmers2023qlora}
Tim Dettmers, Artidoro Pagnoni, Ari Holtzman, and Luke Zettlemoyer. 2023.
\newblock Qlora: Efficient finetuning of quantized llms.
\newblock \emph{arXiv preprint arXiv:2305.14314}.

\bibitem[{Dong et~al.(2023)Dong, Li, Dai, Zheng, Wu, Chang, Sun, Xu, Li, and Sui}]{dong2023survey}
Qingxiu Dong, Lei Li, Damai Dai, Ce~Zheng, Zhiyong Wu, Baobao Chang, Xu~Sun, Jingjing Xu, Lei Li, and Zhifang Sui. 2023.
\newblock \href {http://arxiv.org/abs/2301.00234} {A survey on in-context learning}.

\bibitem[{Dosovitskiy et~al.(2021)Dosovitskiy, Beyer, Kolesnikov, Weissenborn, Zhai, Unterthiner, Dehghani, Minderer, Heigold, Gelly, Uszkoreit, and Houlsby}]{dosovitskiy2021image}
Alexey Dosovitskiy, Lucas Beyer, Alexander Kolesnikov, Dirk Weissenborn, Xiaohua Zhai, Thomas Unterthiner, Mostafa Dehghani, Matthias Minderer, Georg Heigold, Sylvain Gelly, Jakob Uszkoreit, and Neil Houlsby. 2021.
\newblock \href {http://arxiv.org/abs/2010.11929} {An image is worth 16x16 words: Transformers for image recognition at scale}.

\bibitem[{ElSherief et~al.(2021)ElSherief, Ziems, Muchlinski, Anupindi, Seybolt, De~Choudhury, and Yang}]{elsherief-etal-2021-latent}
Mai ElSherief, Caleb Ziems, David Muchlinski, Vaishnavi Anupindi, Jordyn Seybolt, Munmun De~Choudhury, and Diyi Yang. 2021.
\newblock \href {https://aclanthology.org/2021.emnlp-main.29} {Latent hatred: A benchmark for understanding implicit hate speech}.
\newblock In \emph{Proceedings of the 2021 Conference on Empirical Methods in Natural Language Processing}, pages 345--363, Online and Punta Cana, Dominican Republic. Association for Computational Linguistics.

\bibitem[{Ghosh et~al.(2024{\natexlab{a}})Ghosh, Acharya, Jha, Saha, Gaudgaul, Majumdar, Chadha, Jain, Sinha, and Agarwal}]{ghosh2024medsumm}
Akash Ghosh, Arkadeep Acharya, Prince Jha, Sriparna Saha, Aniket Gaudgaul, Rajdeep Majumdar, Aman Chadha, Raghav Jain, Setu Sinha, and Shivani Agarwal. 2024{\natexlab{a}}.
\newblock Medsumm: A multimodal approach to summarizing code-mixed hindi-english clinical queries.
\newblock In \emph{European Conference on Information Retrieval}, pages 106--120. Springer.

\bibitem[{Ghosh et~al.(2024{\natexlab{b}})Ghosh, Acharya, Saha, Jain, and Chadha}]{ghosh2024exploring}
Akash Ghosh, Arkadeep Acharya, Sriparna Saha, Vinija Jain, and Aman Chadha. 2024{\natexlab{b}}.
\newblock Exploring the frontier of vision-language models: A survey of current methodologies and future directions.
\newblock \emph{arXiv preprint arXiv:2404.07214}.

\bibitem[{Girdhar et~al.(2023)Girdhar, El-Nouby, Liu, Singh, Alwala, Joulin, and Misra}]{girdhar2023imagebind}
Rohit Girdhar, Alaaeldin El-Nouby, Zhuang Liu, Mannat Singh, Kalyan~Vasudev Alwala, Armand Joulin, and Ishan Misra. 2023.
\newblock \href {http://arxiv.org/abs/2305.05665} {Imagebind: One embedding space to bind them all}.

\bibitem[{Grootendorst(2022)}]{grootendorst2022bertopic}
Maarten Grootendorst. 2022.
\newblock Bertopic: Neural topic modeling with a class-based tf-idf procedure.
\newblock \emph{arXiv preprint arXiv:2203.05794}.

\bibitem[{He et~al.(2023)He, Ahamad, and Kumar}]{he2023reinforcement}
Bing He, Mustaque Ahamad, and Srijan Kumar. 2023.
\newblock Reinforcement learning-based counter-misinformation response generation: A case study of covid-19 vaccine misinformation.
\newblock In \emph{Proceedings of the ACM Web Conference 2023}.

\bibitem[{Hee et~al.(2023)Hee, Chong, and Lee}]{hee2023decoding}
Ming~Shan Hee, Wen-Haw Chong, and Roy Ka-Wei Lee. 2023.
\newblock Decoding the underlying meaning of multimodal hateful memes.
\newblock \emph{arXiv preprint arXiv:2305.17678}.

\bibitem[{Holtzman et~al.(2020)Holtzman, Buys, Du, Forbes, and Choi}]{holtzman2020curious}
Ari Holtzman, Jan Buys, Li~Du, Maxwell Forbes, and Yejin Choi. 2020.
\newblock \href {http://arxiv.org/abs/1904.09751} {The curious case of neural text degeneration}.

\bibitem[{Houlsby et~al.(2019)Houlsby, Giurgiu, Jastrzebski, Morrone, de~Laroussilhe, Gesmundo, Attariyan, and Gelly}]{houlsby2019parameterefficient}
Neil Houlsby, Andrei Giurgiu, Stanislaw Jastrzebski, Bruna Morrone, Quentin de~Laroussilhe, Andrea Gesmundo, Mona Attariyan, and Sylvain Gelly. 2019.
\newblock \href {http://arxiv.org/abs/1902.00751} {Parameter-efficient transfer learning for nlp}.

\bibitem[{Hu et~al.(2021)Hu, Shen, Wallis, Allen-Zhu, Li, Wang, Wang, and Chen}]{hu2021lora}
Edward~J Hu, Yelong Shen, Phillip Wallis, Zeyuan Allen-Zhu, Yuanzhi Li, Shean Wang, Lu~Wang, and Weizhu Chen. 2021.
\newblock Lora: Low-rank adaptation of large language models.
\newblock \emph{arXiv preprint arXiv:2106.09685}.

\bibitem[{Hua et~al.(2019)Hua, Hu, and Wang}]{hua2019argument}
Xinyu Hua, Zhe Hu, and Lu~Wang. 2019.
\newblock Argument generation with retrieval, planning, and realization.
\newblock \emph{arXiv preprint arXiv:1906.03717}.

\bibitem[{Hwang and Shwartz(2023)}]{hwang2023memecap}
EunJeong Hwang and Vered Shwartz. 2023.
\newblock \href {http://arxiv.org/abs/2305.13703} {Memecap: A dataset for captioning and interpreting memes}.

\bibitem[{Jain et~al.(2023)Jain, Maity, Jha, and Saha}]{DBLP:conf/ijcnn/JainMJS23}
Raghav Jain, Krishanu Maity, Prince Jha, and Sriparna Saha. 2023.
\newblock \href {https://doi.org/10.1109/IJCNN54540.2023.10191363} {Generative models vs discriminative models: Which performs better in detecting cyberbullying in memes?}
\newblock In \emph{International Joint Conference on Neural Networks, {IJCNN} 2023, Gold Coast, Australia, June 18-23, 2023}, pages 1--8. {IEEE}.

\bibitem[{Jha et~al.(2024)Jha, Maity, Jain, Verma, Saha, and Bhattacharyya}]{DBLP:conf/eacl/JhaMJVSB24}
Prince Jha, Krishanu Maity, Raghav Jain, Apoorv Verma, Sriparna Saha, and Pushpak Bhattacharyya. 2024.
\newblock \href {https://aclanthology.org/2024.eacl-long.56} {Meme-ingful analysis: Enhanced understanding of cyberbullying in memes through multimodal explanations}.
\newblock In \emph{Proceedings of the 18th Conference of the European Chapter of the Association for Computational Linguistics, {EACL} 2024 - Volume 1: Long Papers, St. Julian's, Malta, March 17-22, 2024}, pages 930--943. Association for Computational Linguistics.

\bibitem[{Kiela et~al.(2020)Kiela, Firooz, Mohan, Goswami, Singh, Ringshia, and Testuggine}]{kiela2020hateful}
Douwe Kiela, Hamed Firooz, Aravind Mohan, Vedanuj Goswami, Amanpreet Singh, Pratik Ringshia, and Davide Testuggine. 2020.
\newblock The hateful memes challenge: Detecting hate speech in multimodal memes.
\newblock \emph{Advances in Neural Information Processing Systems}, 33:2611--2624.

\bibitem[{Kojima et~al.(2023)Kojima, Gu, Reid, Matsuo, and Iwasawa}]{kojima2023large}
Takeshi Kojima, Shixiang~Shane Gu, Machel Reid, Yutaka Matsuo, and Yusuke Iwasawa. 2023.
\newblock \href {http://arxiv.org/abs/2205.11916} {Large language models are zero-shot reasoners}.

\bibitem[{Lin(2004)}]{lin2004rouge}
Chin-Yew Lin. 2004.
\newblock Rouge: A package for automatic evaluation of summaries.
\newblock In \emph{Text summarization branches out}, pages 74--81.

\bibitem[{Maity et~al.(2023)Maity, Jain, Jha, Saha, and Bhattacharyya}]{DBLP:conf/emnlp/MaityJJ0B23}
Krishanu Maity, Raghav Jain, Prince Jha, Sriparna Saha, and Pushpak Bhattacharyya. 2023.
\newblock \href {https://doi.org/10.18653/V1/2023.EMNLP-MAIN.1035} {Genex: {A} commonsense-aware unified generative framework for explainable cyberbullying detection}.
\newblock In \emph{Proceedings of the 2023 Conference on Empirical Methods in Natural Language Processing, {EMNLP} 2023, Singapore, December 6-10, 2023}, pages 16632--16645. Association for Computational Linguistics.

\bibitem[{Maity et~al.(2022)Maity, Jha, Saha, and Bhattacharyya}]{10.1145/3477495.3531925}
Krishanu Maity, Prince Jha, Sriparna Saha, and Pushpak Bhattacharyya. 2022.
\newblock \href {https://doi.org/10.1145/3477495.3531925} {A multitask framework for sentiment, emotion and sarcasm aware cyberbullying detection from multi-modal code-mixed memes}.
\newblock In \emph{Proceedings of the 45th International ACM SIGIR Conference on Research and Development in Information Retrieval}, SIGIR '22, page 1739–1749, New York, NY, USA. Association for Computing Machinery.

\bibitem[{Maity et~al.(2024)Maity, Poornash, Bhattacharya, Phosit, Kongsamlit, Saha, and Pasupa}]{10494986}
Krishanu Maity, A.~S. Poornash, Shaubhik Bhattacharya, Salisa Phosit, Sawarod Kongsamlit, Sriparna Saha, and Kitsuchart Pasupa. 2024.
\newblock \href {https://doi.org/10.1109/TCSS.2024.3376958} {Hatethaisent: Sentiment-aided hate speech detection in thai language}.
\newblock \emph{IEEE Transactions on Computational Social Systems}, pages 1--14.

\bibitem[{Mangrulkar et~al.(2022)Mangrulkar, Gugger, Debut, Belkada, and Paul}]{peft}
Sourab Mangrulkar, Sylvain Gugger, Lysandre Debut, Younes Belkada, and Sayak Paul. 2022.
\newblock Peft: State-of-the-art parameter-efficient fine-tuning methods.
\newblock \url{https://github.com/huggingface/peft}.

\bibitem[{Mathew et~al.(2018)Mathew, Kumar, Goyal, Mukherjee et~al.}]{mathew2018analyzing}
Binny Mathew, Navish Kumar, Pawan Goyal, Animesh Mukherjee, et~al. 2018.
\newblock Analyzing the hate and counter speech accounts on twitter.
\newblock \emph{arXiv preprint arXiv:1812.02712}.

\bibitem[{Mathias et~al.(2021)Mathias, Nie, Davani, Kiela, Prabhakaran, Vidgen, and Waseem}]{mathias2021findings}
Lambert Mathias, Shaoliang Nie, Aida~Mostafazadeh Davani, Douwe Kiela, Vinodkumar Prabhakaran, Bertie Vidgen, and Zeerak Waseem. 2021.
\newblock Findings of the woah 5 shared task on fine grained hateful memes detection.
\newblock In \emph{Proceedings of the 5th Workshop on Online Abuse and Harms (WOAH 2021)}, pages 201--206.

\bibitem[{Ohlheiser(2016)}]{ohlheiser2016banned}
Abby Ohlheiser. 2016.
\newblock Banned from twitter? this site promises you can say whatever you want.
\newblock \emph{Washington Post}, 29.

\bibitem[{Papineni et~al.(2002)Papineni, Roukos, Ward, and Zhu}]{papineni2002bleu}
Kishore Papineni, Salim Roukos, Todd Ward, and Wei-Jing Zhu. 2002.
\newblock Bleu: a method for automatic evaluation of machine translation.
\newblock In \emph{Proceedings of the 40th annual meeting of the Association for Computational Linguistics}, pages 311--318.

\bibitem[{Pramanick et~al.(2021{\natexlab{a}})Pramanick, Dimitrov, Mukherjee, Sharma, Akhtar, Nakov, Chakraborty et~al.}]{pramanick2021detecting}
Shraman Pramanick, Dimitar Dimitrov, Rituparna Mukherjee, Shivam Sharma, Md~Akhtar, Preslav Nakov, Tanmoy Chakraborty, et~al. 2021{\natexlab{a}}.
\newblock Detecting harmful memes and their targets.
\newblock \emph{arXiv preprint arXiv:2110.00413}.

\bibitem[{Pramanick et~al.(2021{\natexlab{b}})Pramanick, Sharma, Dimitrov, Akhtar, Nakov, and Chakraborty}]{pramanick2021momenta}
Shraman Pramanick, Shivam Sharma, Dimitar Dimitrov, Md~Shad Akhtar, Preslav Nakov, and Tanmoy Chakraborty. 2021{\natexlab{b}}.
\newblock Momenta: A multimodal framework for detecting harmful memes and their targets.
\newblock \emph{arXiv preprint arXiv:2109.05184}.

\bibitem[{Qian et~al.(2019)Qian, Bethke, Liu, Belding, and Wang}]{qian2019benchmark}
Jing Qian, Anna Bethke, Yinyin Liu, Elizabeth Belding, and William~Yang Wang. 2019.
\newblock A benchmark dataset for learning to intervene in online hate speech.
\newblock \emph{arXiv preprint arXiv:1909.04251}.

\bibitem[{Schieb and Preuss(2016)}]{schieb2016governing}
Carla Schieb and Mike Preuss. 2016.
\newblock Governing hate speech by means of counterspeech on facebook.
\newblock In \emph{66th ica annual conference, at fukuoka, japan}, pages 1--23.

\bibitem[{Sharma et~al.(2023)Sharma, Arora, Akhtar, Chakraborty et~al.}]{sharma2023memex}
Shivam Sharma, Udit Arora, Md~Shad Akhtar, Tanmoy Chakraborty, et~al. 2023.
\newblock Memex: Detecting explanatory evidence for memes via knowledge-enriched contextualization.
\newblock \emph{arXiv preprint arXiv:2305.15913}.

\bibitem[{Shi et~al.(2023)Shi, Chen, Misra, Scales, Dohan, Chi, Schärli, and Zhou}]{shi2023large}
Freda Shi, Xinyun Chen, Kanishka Misra, Nathan Scales, David Dohan, Ed~Chi, Nathanael Schärli, and Denny Zhou. 2023.
\newblock \href {http://arxiv.org/abs/2302.00093} {Large language models can be easily distracted by irrelevant context}.

\bibitem[{Smith et~al.(2008)Smith, Mahdavi, Carvalho, Fisher, Russell, and Tippett}]{smith2008cyberbullying}
Peter~K Smith, Jess Mahdavi, Manuel Carvalho, Sonja Fisher, Shanette Russell, and Neil Tippett. 2008.
\newblock Cyberbullying: Its nature and impact in secondary school pupils.
\newblock \emph{Journal of child psychology and psychiatry}, 49(4):376--385.

\bibitem[{Touvron et~al.(2023)Touvron, Lavril, Izacard, Martinet, Lachaux, Lacroix, Rozière, Goyal, Hambro, Azhar, Rodriguez, Joulin, Grave, and Lample}]{touvron2023llama}
Hugo Touvron, Thibaut Lavril, Gautier Izacard, Xavier Martinet, Marie-Anne Lachaux, Timothée Lacroix, Baptiste Rozière, Naman Goyal, Eric Hambro, Faisal Azhar, Aurelien Rodriguez, Armand Joulin, Edouard Grave, and Guillaume Lample. 2023.
\newblock \href {http://arxiv.org/abs/2302.13971} {Llama: Open and efficient foundation language models}.

\bibitem[{Veselovsky et~al.(2023)Veselovsky, Ribeiro, and West}]{veselovsky2023artificial}
Veniamin Veselovsky, Manoel~Horta Ribeiro, and Robert West. 2023.
\newblock Artificial artificial artificial intelligence: Crowd workers widely use large language models for text production tasks.
\newblock \emph{arXiv preprint arXiv:2306.07899}.

\bibitem[{Wang et~al.(2023)Wang, Hee, Awal, Choo, and Lee}]{wang2023evaluating}
Han Wang, Ming~Shan Hee, Md~Rabiul Awal, Kenny Tsu~Wei Choo, and Roy Ka-Wei Lee. 2023.
\newblock Evaluating gpt-3 generated explanations for hateful content moderation.
\newblock \emph{arXiv preprint arXiv:2305.17680}.

\bibitem[{Wright et~al.(2017)Wright, Ruths, Dillon, Saleem, and Benesch}]{wright2017vectors}
Lucas Wright, Derek Ruths, Kelly~P Dillon, Haji~Mohammad Saleem, and Susan Benesch. 2017.
\newblock Vectors for counterspeech on twitter.
\newblock In \emph{Proceedings of the first workshop on abusive language online}, pages 57--62.

\bibitem[{Ybarra et~al.(2006)Ybarra, Mitchell, Wolak, and Finkelhor}]{ybarra2006examining}
Michele~L Ybarra, Kimberly~J Mitchell, Janis Wolak, and David Finkelhor. 2006.
\newblock Examining characteristics and associated distress related to internet harassment: findings from the second youth internet safety survey.
\newblock \emph{Pediatrics}, 118(4):e1169--e1177.

\bibitem[{Yuan et~al.(2023)Yuan, Xue, Wang, Liu, Zhao, and Wang}]{yuan2023artgpt4}
Zhengqing Yuan, Huiwen Xue, Xinyi Wang, Yongming Liu, Zhuanzhe Zhao, and Kun Wang. 2023.
\newblock \href {http://arxiv.org/abs/2305.07490} {Artgpt-4: Artistic vision-language understanding with adapter-enhanced minigpt-4}.

\bibitem[{Zhang et~al.(2023)Zhang, Zhang, Xu, and Tao}]{zhang2023vision}
Qiming Zhang, Jing Zhang, Yufei Xu, and Dacheng Tao. 2023.
\newblock Vision transformer with quadrangle attention.
\newblock \emph{arXiv preprint arXiv:2303.15105}.

\bibitem[{Zhang* et~al.(2020)Zhang*, Kishore*, Wu*, Weinberger, and Artzi}]{bert-score}
Tianyi Zhang*, Varsha Kishore*, Felix Wu*, Kilian~Q. Weinberger, and Yoav Artzi. 2020.
\newblock \href {https://openreview.net/forum?id=SkeHuCVFDr} {Bertscore: Evaluating text generation with bert}.
\newblock In \emph{International Conference on Learning Representations}.

\bibitem[{Zhu et~al.(2023)Zhu, Chen, Shen, Li, and Elhoseiny}]{zhu2023minigpt}
Deyao Zhu, Jun Chen, Xiaoqian Shen, Xiang Li, and Mohamed Elhoseiny. 2023.
\newblock Minigpt-4: Enhancing vision-language understanding with advanced large language models.
\newblock \emph{arXiv preprint arXiv:2304.10592}.

\end{thebibliography}
\bibliographystyle{acl_natbib}

\newpage

\renewcommand{\thesubsection}{\Alph{section}.\arabic{subsection}}
\renewcommand{\thesection}{\Alph{section}}
\setcounter{section}{0}

\section{Appendix}
\label{sec:appendix}

This section provides supplementary material in the form of FAQs, ethical considerations, additional results, implementation details, etc., to bolster the reader's understanding of the concepts presented in this work.

\section*{Frequently Asked Questions (FAQs)}\label{sec:FAQs}

\begin{itemize}
[leftmargin=5mm]
\setlength\itemsep{1.5em}
    \item[\ding{93}] {\fontfamily{lmss} \selectfont \textbf{What was the reasoning behind producing only 1000 datasets for the meme intervention task?}}
    \vspace{-2mm}
    \begin{description}
    \item[\ding{224}] We chose to generate a smaller dataset, specifically of 1000 meme interventions, due to the complex and labor-intensive nature of the task. The creation process of a meme intervention demands a deep understanding of the meme and the identification of any toxic elements within it, followed by the generation of the intervention based on specific guidelines. We refrained from using crowdsourcing platforms due to concerns about the potential compromise in data quality. This concern is substantiated by recent evidence \cite{veselovsky2023artificial} showing that even crowdsourced workers rely on tools like GPT3.5-Turbo for their tasks. Our approach prioritized quality over quantity, with an aim to create a meticulously refined and high-quality dataset. While this method ensured superior dataset quality, it was significantly more time-consuming.
    \end{description}

    \item[\ding{93}] {\fontfamily{lmss} \selectfont \textbf{What was the rationale behind selecting these specific llms?}}
    \vspace{-2mm}
    \begin{description}
    \item[\ding{224}] Our goal in using a wide range of Large Language Models (LLMs) for our MemeGuard framework was to incorporate diversity, from autoregressive decoder-only models to encoder-decoder models and instruction-tuned models. We chose GPT3.5-Turbo because it is one of the highest-performing LLMs currently available. However, due to its proprietary nature, we supplemented it with recently launched instruction-tuned LLMs, namely RedPajama and Dolly. One distinct feature of these models is that GPT3.5-Turbo has been trained using the Reinforcement Learning from Human Feedback (RLHF) strategy, whereas RedPajama and Dolly have employed supervised instruction fine-tuning. We also included FLAN-T5 in our selection due to its unique encoder-decoder architecture, contrasting with the other instruction-tuned LLMs. The inclusion of LLama demonstrates the performance of an autoregressive decoder-only model.
    \end{description}

    \item[\ding{93}] {\fontfamily{lmss} \selectfont \textbf{Could this methodology be implemented in other domains, like misinformation?}}
    \vspace{-2mm}
    \begin{description}
    \item[\ding{224}] Absolutely, the approach we've developed isn't exclusive to tackling toxic memes. It holds potential for application in other areas like misinformation and disinformation. Essentially, all components would remain the same; the only change required would be to substitute VLMeme with a domain-specific Vision and Language Model (VLM), or even a general-purpose one. We haven't experimented with this in other domains as, to the best of our knowledge, there aren't any other multimodal intervention datasets currently available for such applications.
    \end{description}
    \item[\ding{93}] {\fontfamily{lmss} \selectfont \textbf{Could this methodology used for developing VLMeme be utilized for other Vision and Language Models (VLMs)?}}
    \vspace{-2mm}
    \begin{description}
    \item[\ding{224}] Indeed, the image adapter fine-tuning strategy deployed in this study can be extrapolated to other Vision and Language Models (VLMs). Nonetheless, our empirical examination was confined to the application on MiniGPT4. We wish to emphasize that there is a panoply of other effective fine-tuning strategies \cite{chavan2023one,peft} that have not been explored within the confines of this research. Future endeavors will involve the investigation and identification of the optimal VLM and fine-tuning strategy for the complex task of meme comprehension.
    \end{description}
    \item[\ding{93}] {\fontfamily{lmss} \selectfont \textbf{Have the BERTScores been rescaled according to the guidelines outlined in \cite{rescale} for clarification?}}
    \vspace{-2mm}
    \begin{description}
    \item[\ding{224}]  We did not rescale the BERTScore concerning its empirical lower bound 'b' as a baseline. Our reported average F1 BERTScore aligns with the documentation provided at \href{https://huggingface.co/spaces/evaluate-metric/bertscore}{huggingface} .
    \end{description}

     \item[\ding{93}] {\fontfamily{lmss} \selectfont \textbf{Have you considered utilising various adapters such as LoRA~\cite{hu2021lora}, QLoRA~\cite{dettmers2023qlora} in your experiments?}}
    \vspace{-2mm}
    \begin{description}
    \item[\ding{224}]  
    Thank you for suggesting LoRA and QLoRA as potential adapters for our experiments. However, in this specific study, we didn't integrate these adapters. The primary focus of our paper lies in proposing interventions for combating the toxicity of cyberbullying memes. Nevertheless, we acknowledge the potential advantages of such an approach and will indeed consider exploring it in our future research endeavors.
    \end{description}

      \item[\ding{93}] {\fontfamily{lmss} \selectfont \textbf{ Could you provide clarification on the development process of the prompts used in the study ?}}
    \vspace{-2mm}
    \begin{description}
    \item[\ding{224}]  
   To ensure the effectiveness and relevance of our prompts, we conducted a qualitative analysis on a diverse yet small sample set. This preliminary analysis allowed us to gauge the responsiveness and efficacy of different prompts in a controlled setting. Based on the insights gathered from this analysis, we were able to refine and finalize the prompts used in our experiments. This method, while not directly derived from similar studies, was crucial in ensuring that our prompts were empirically sound and tailored to the specific objectives of our research.

    \end{description}

       \item[\ding{93}] {\fontfamily{lmss} \selectfont \textbf{ Could you please provide an analysis or discussion regarding why GPT3.5-Tubro's scores consistently outperform other models?}}
    \vspace{-2mm}
    \begin{description}
    \item[\ding{224}]  
     The reason why GPT3.5-Turbo performed well: GPT3.5-Turbo training includes reinforcement learning from human feedback (RLHF), which could fine-tune its outputs based on human preferences and evaluations, leading to more human-like and contextually appropriate responses.

    \end{description}

    \item[\ding{93}] {\fontfamily{lmss} \selectfont \textbf{How is the OCR done?}}
    \vspace{-2mm}
    \begin{description}
    \item[\ding{224}]  
  OCR was extracted using the Google Cloud Vision API. We did not perform the OCR extraction ourselves. In the MultiBully dataset, OCR information was provided. However, the authors of the dataset mentioned in their paper that they utilized the Google Cloud Vision API\footnote{\url{https://cloud.google.com/vision/docs/ocr}} for the extraction process.

    \end{description}

        \item[\ding{93}] {\fontfamily{lmss} \selectfont \textbf{Baselines are designed by authors, while some variations of existing textual intervention methods are also expected.}}
    \vspace{-2mm}
    \begin{description}
    \item[\ding{224}] 
In this paper, we introduce MemeGuard, which employs a meme-aligned Vision-Language Model (VLMeme) to generate contextual information about the meme, subsequently used for the final intervention generation. We also prompt various Language Models (LLMs), including Dolly, LLaMA, RedPajama, FLAN-T5, and GPT3.5 Turbo, in a unimodal setting (i.e., OCR only) to assess their ability to generate interventions in a zero-shot manner. Our proposed method, MemeGuard does not require any training on an annotated intervention dataset for automatic intervention generation. Existing text-based approaches for intervention generation necessitate training over an intervention-annotated corpus and are unsuitable for prompting, rendering them incomparable with our framework. This discrepancy does not align with the goals and contributions of this paper.

    \end{description}
    
\end{itemize}

\section*{Ethical Considerations}\label{sec: ethics}
\textbf{Reproducibility: }
We have provided comprehensive details of our experimental setups, including hyperparameters and evaluation metrics, in Appendix~\ref{sec:eval}, aiming to facilitate reproducibility. Upon acceptance of the paper, we will make our code and dataset publicly available.\\
\textbf{User Privacy: }The information depicted or utilized does not contain any personal information.\\
\textbf{Annotation: }Repetitive consumption of online abuse could distress mental health conditions \cite{ybarra2006examining}. Therefore, we advised annotators to take periodic breaks and not do the annotations in one sitting. Besides, we had weekly meetings with them to ensure the annotators did not have any adverse effect on their mental health. \\
\textbf{Biases: }We instructed annotators to annotate the posts without considering any specific demographic, racial, religious, or other factors. However, memes can be subjective, leading to inherent biases in our gold standard dataset. Any biases identified within our dataset are unintended, and we have no intention to harm any individual or group\\
\textbf{Misuse Potential: }
The implementation of interventions in hate speech carries the potential for misuse, including the suppression of free speech and the targeting of specific individuals or groups based on personal or ideological agendas. The mechanisms and algorithms utilized to detect and moderate hate speech may lack transparency, leading to difficulties for users in comprehending why their content was flagged or removed. Moreover, when the criteria for identifying hate speech are vague or overly broad, there exists a risk that individuals expressing controversial yet lawful viewpoints may face unjust targeting or silencing. This concern can be addressed through the establishment of lawful regulations to define hate speech and the inclusion of human intervention to ensure fair moderation.\\
\textbf{Intended Use: } We have annotated the ICMM dataset for research purposes, adhering to the usage policies set forth by various sources/platforms. We follow similar principles in the entirety of its usage as well. The distribution of this dataset will be limited to research purposes only, without granting a license for commercial use. We firmly believe that it is a valuable resource when utilized appropriately\\
\textbf{Review Board:} The institute's review board has approved the data collection and annotation protocol.

\subsection{Experimental Setup}
\label{sec:eval}
\textbf{Evaluation Metrics:} 
We utilize automatic evaluations as a means to assess the quality of the generated interventions. To measure the similarity between the generated text and the ground truth, we employ ROUGE~\cite{lin2004rouge} and BLEU~\cite{papineni2002bleu} scores, which analyze the overlap of N-grams. Additionally, we use BERTScore to evaluate the semantic similarity between the generated interventions and the reference.

In order to observe the combined impact of ROUGE and BLEU, we calculate the average BLEU score and the harmonic mean of ROUGE-L and the average BLEU score as described in \citet{hee2023decoding}. These metrics provide a comprehensive view of both the lexical and semantic aspects of the generated interventions.
Furthermore, we conduct a human evaluation on the most effective models. For the human evaluation, we enlist human evaluators who assess the generated interventions based on four aspects: fluency, adequacy, persuasiveness, and informativeness. The evaluators are instructed to rate the generated interventions using the Likert scales described in section~\ref{sec:ann-quality}. Table \ref{LLMs} presents different details about LLMs used in this study.

\subsection{Hyperparameters}
\label{hype}
\begin{figure}[hbt]
	\centering
	\includegraphics[height = 5 cm, width = 7cm]{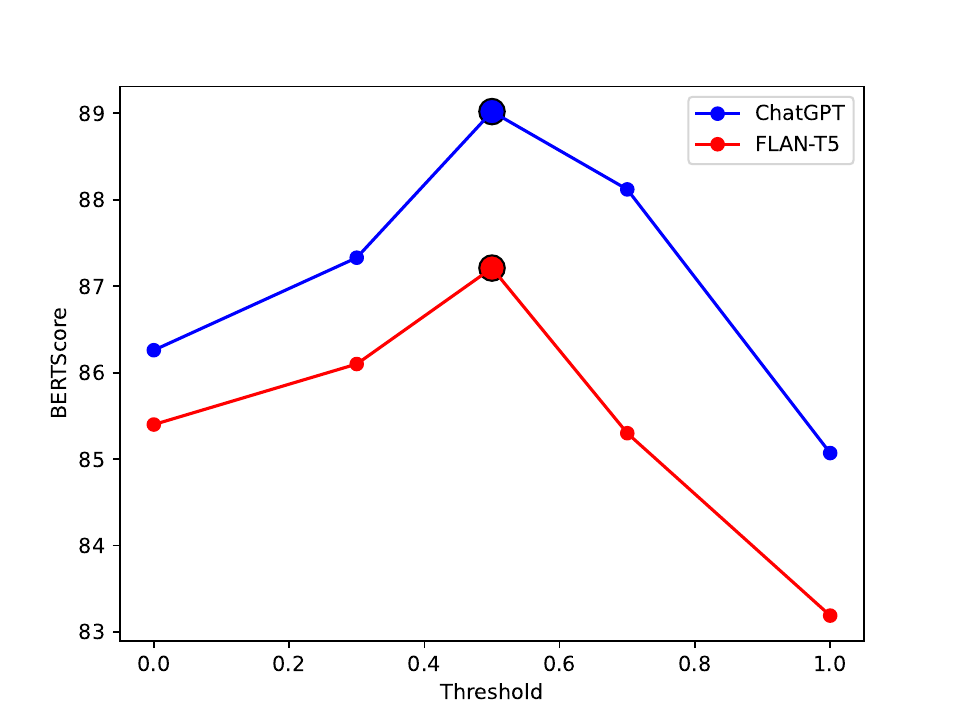}
	\caption {BERTScore variation for GPT3.5-Turbo and FLAN-T5 with different Threshold values}
	\label{plot}
\end{figure}

\textbf{Threshold $Th$: }We conducted an extensive study to determine the ideal similarity threshold value, denoted as $Th$, for the MKS module. This investigation involved varying the $Th$ between 0 and 1 for both FLAN-T5 and GPT3.5-Turbo. As indicated in figure \ref{plot}, it was found that these models both achieved their highest BERTScores at $Th=0.5$. Consequently, we decided to establish the $Th$ at 0.5.\\ 
We configured the other hyperparameters in the following manner: a temperature setting of 0.5, a $top\_p$ value of 0.2, and a $top\_k$ value of 50.
\begin{table}[!b]
\centering
\scalebox{0.7}{%
\begin{tabular}{l|c|l|c}
\hline
\textbf{Language Model} & \textbf{Params} & \textbf{Architecture}                & \textbf{Type} \\ \hline
Dolly          & 3B     & Autoregressive Decoder Only & Base \\ \hline
LLaMA          & 7B     & Autoregressive Decoder Only & Base \\ \hline
RedPajama      & 3B     & Autoregressive Decoder Only & Base \\ \hline
FLAN-T5           & 780M   & Encoder- Decoder            & SIFT \\ \hline
GPT3.5-Turbo        & -      & -                           & RLHF \\ \hline
\end{tabular}%
}
\caption{Characterstics of Different LLMs used in this study. Base denotes standard pre-training strategies, SIFT means Supervised Instruction Fine Tuning and RLHF means Reinforcement Learning from Human Feedback}
\label{LLMs}
\end{table}

\subsection{Performance of VLMeme}
\label{vlmsec}
We evaluated the efficacy of our fine-tuned, meme-aligned VLMeme on a meme description task by comparing its performance against multiple baselines, as presented in the MemeCap dataset paper \cite{hwang2023memecap}. The findings, as illustrated in table \ref{vlm}, reveal that VLMeme surpasses all other baselines across all measures, underscoring its superior comprehension of memes compared to the standard MiniGPT-4. 
\begin{table}[]
\centering
\scalebox{0.7}{%
\begin{tabular}{cc|ccc|}
\cline{2-5}
\multicolumn{1}{c|}{}                                    & \textbf{\acb{Prompting}}     & \textbf{BLEU-4} & \textbf{ROUGE-L} & \textbf{BERT-F1} \\ \hline
\multicolumn{1}{|c|}{\multirow{4}{*}{\rotatebox[origin=c]{90}{\textbf{MiniGPT4}}}} & {Zero-shot} & 12.46           & 31.44            & 68.62            \\
\multicolumn{1}{|c|}{} & {Zero-shot CoT}  & 12.57 & 31.7  & 68.45 \\
\multicolumn{1}{|c|}{} & {Fine-tuned}     & 7.5   & 27.88 & 65.47 \\
\multicolumn{1}{|c|}{} & {Fine-tuned CoT} & 7.25  & 26.68 & 65.86 \\ \hline \hline
\multicolumn{2}{|c|}{\textbf{VLMeme}}                     & \textbf{13.31}      & \textbf{33.01}      &  \textbf{79.82}     \\ \hline
\end{tabular}%
}
\caption{Performance of VLMeme across different metrics on MemeCap dataset.}
\label{vlm}
\end{table}


\begin{figure*}[hbt]
	\centering
	\includegraphics[height = 10 cm, width = 17cm]{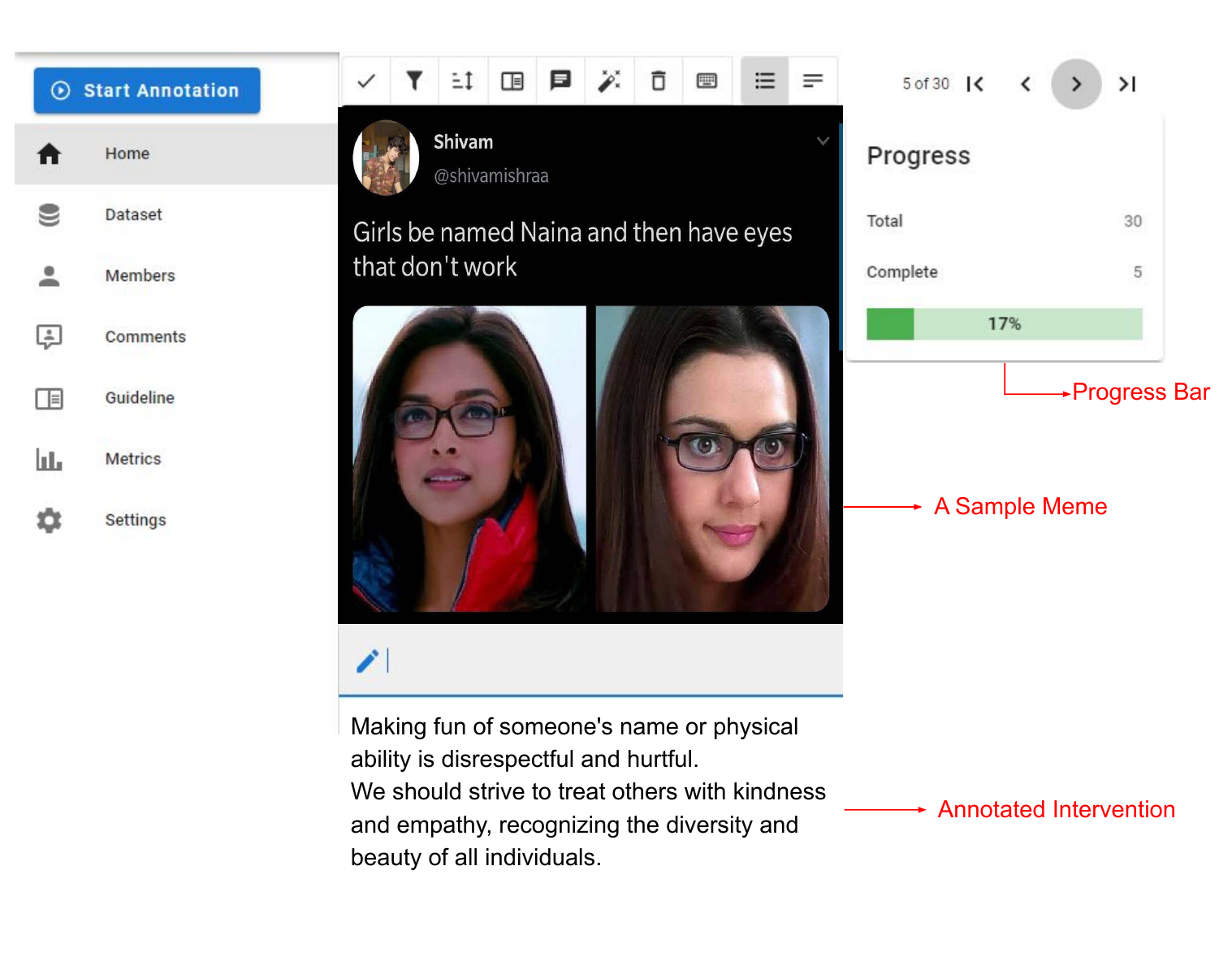}
 \vspace{-5mm}
	\caption {A screenshot of the annotation platform toolkit.}
	\label{fig:ann_toolkit}
\end{figure*} 
\subsection{Annotation Guidelines}
\label{AnnGd}
We follow cyberbullying definition by \citet{smith2008cyberbullying} for our annotation process. In order to help and guide our annotators, we provide them with several examples of memes with expert annotated intervention. Motivated by argument generation style of text planning decoder in \citet{hua2019argument}, we write an intervention to each meme in two sentences: \\
\textbf{(1) Interventive Content:} The sentence which delivers critical ideas for mitigation of cyberbullying based on toxic information related to gender, race, religion, nationality, physical ability, mental ability, stereotype, societal biases, etc... present in the meme, e.g. "While bestiality is unethical and warrants condemnation, it is derogatory to defame entire country based on actions of a limited group promoting it."\\
\textbf{(2) Interventive Filler:} The sentence which contains a general statement supporting the interventive content, e.g. "We should strive to use language that is respectful and appropriate in all situations."
\label{sec:guide}

\subsection{Daywise Schedule}
\label{Days}

\begin{itemize}
 \item \textbf{Day 1 and Day 4: } Each annotator was assigned to annotate interventions for 30 memes. They were instructed to annotate 10 memes per batch within one hour, followed by a mandatory break of 20 minutes (cf.  Section~\ref{sec: ethics}).
\item \textbf{Day  2 and Day 5: } Each annotator was assigned the task of evaluating intervention annotations provided by other annotators, assessing them based on fluency, adequacy, informativeness, and persuasiveness.
\item \textbf{Day 3: } We arrange meetings with the annotators to ensure that their mental well-being is not adversely affected during the annotation process (cf. Section~\ref{sec: ethics}).
\end{itemize}

\label{sec:schedule}

\vspace{-3mm}
\subsection{Annotation Quality}
\label{quality}
We assess the quality of annotations based on the following criteria as described in \citet{wang2023evaluating}:\\
\textbf{(1) Fluency:} Rate the structural and grammatical correctness of the interventions using a 5-point Likert scale. 1: represents interventions that are unreadable due to excessive grammatical errors; 5:
represents well-written interventions with no grammatical errors.\\
\textbf{(2) Adequacy: } Rate the adequacy of the intervention using a 5-point Likert scale. 1: interventions misinterpret the implicit stereotypes; 
 5: interventions accurately reflect the implicit stereotypes.\\
 \textbf{(3) Informativeness: } Rate if interventions provide additional background information using a 5-point Likert scale. 1: interventions that are not informative; 5: very informative interventions.\\
 \textbf{(4) Persuasiveness: } Rate the persuasiveness of the intervention using a 5-point Likert scale. 1: interventions that are not persuasive; 5: very persuasive interventions.

\label{sec:ann-quality}

\subsection{Annotation Platform Toolkit}
\label{plat}
\begin{figure*}[hbt]
	\centering
	\includegraphics[height = 10 cm, width = 16cm]{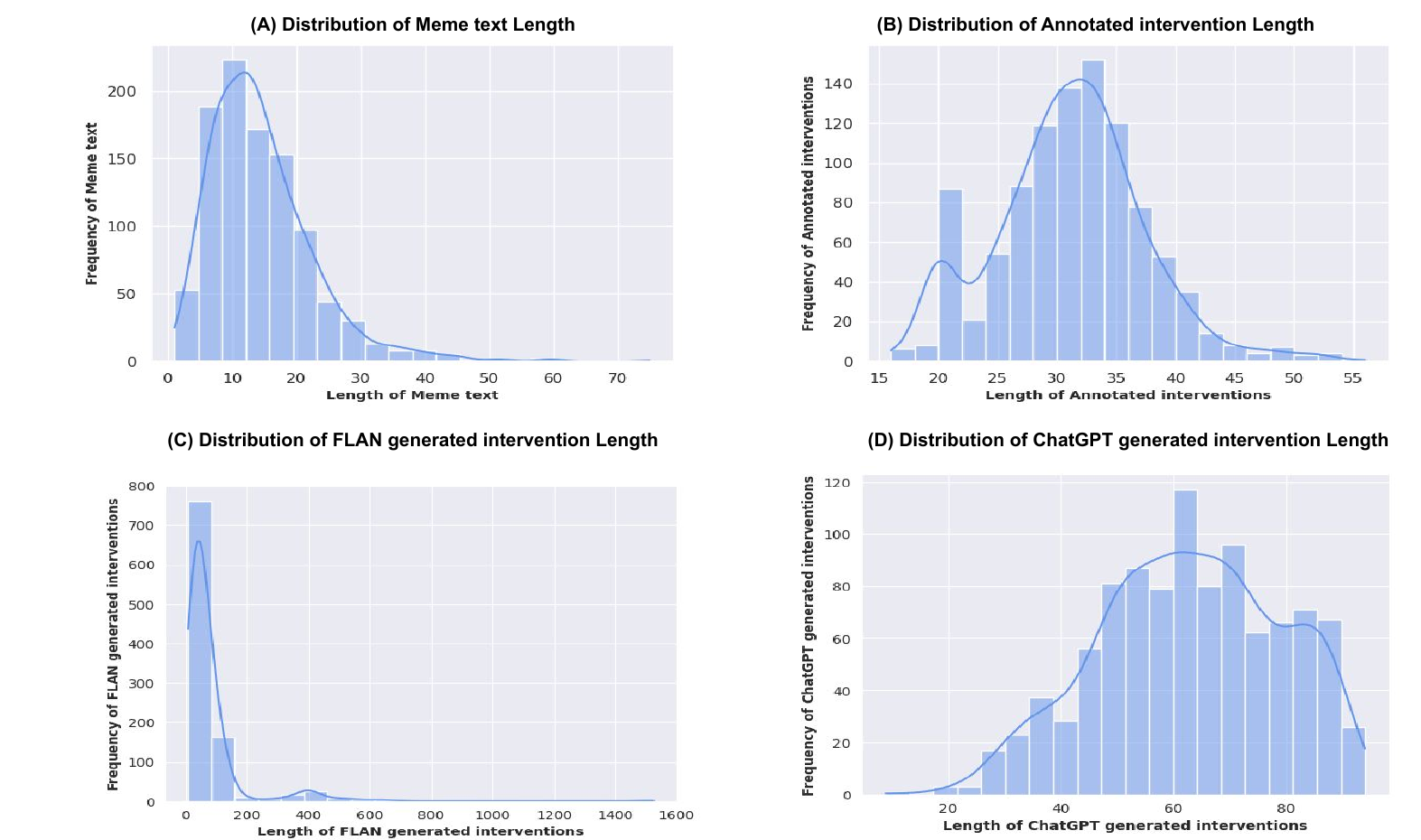}
	\caption {Distribution of Length for Meme Text, Human-annotated interventions, FLAN-generated interventions and GPT3.5-Turbo generated interventions.}
	\label{fig:length_dist}
\end{figure*}
We utilized the open-source platform Docanno\footnote{\url{https://github.com/doccano/doccano}}, which was deployed on a Heroku instance, for our annotation process. Figure~\ref{fig:ann_toolkit} displays a screenshot of the annotation platform toolkit. In the top-left corner of the figure~~\ref{fig:ann_toolkit}, various features are available, including Dataset, which allows for importing the dataset for annotation; Members, used to assign annotators their roles in the project; Guidelines, enabling the sharing of annotation guidelines with all annotators; and Metrics, which facilitates the evaluation of annotated data based on various metrics.

To initiate our annotation process, we began by importing the ICMM dataset. Annotators were able to provide interventions for each meme in the associated text box. Additionally, a progress bar located in the rightmost section of the annotation platform toolkit allowed annotators to track their progress.


\subsection{Annotated v/s Generated Interventions: Length Analysis}
\label{LA}

Figure~\ref{fig:length_dist} illustrates the distribution of meme text length, annotated intervention length, FLAN-generated intervention length, and GPT3.5-Turbo-generated intervention length. In Figure~\ref{fig:length_dist}A, it is evident that the length of meme text varies approximately between 0 and 70 characters. Notably, the interventions generated by FLAN~\ref{fig:length_dist}C exhibit considerably greater length compared to both the GPT3.5-Turbo-generated interventions ~\ref{fig:length_dist}D and the human-annotated interventions~\ref{fig:length_dist}B. Most FLAN-generated interventions fall within the range of 0 to 200 characters, although there are a few interventions with lengths surpassing 200 characters. Conversely, the length distribution of GPT3.5-Turbo-generated interventions and human-annotated interventions follows a more normal distribution pattern. The GPT3.5-Turbo-generated interventions exhibit a scarcity of interventions with a length below approximately 20 characters or exceeding approximately 90 characters. Similarly, the human-annotated interventions display a limited number of interventions shorter than approximately 15 characters or longer than approximately 55 characters.

\subsection{Annotated v/s Generated Interventions: Topical Analysis}
\label{TA}
\begin{figure*}[hbt]
	\centering
 \includegraphics[height = 17cm, width = 17cm]{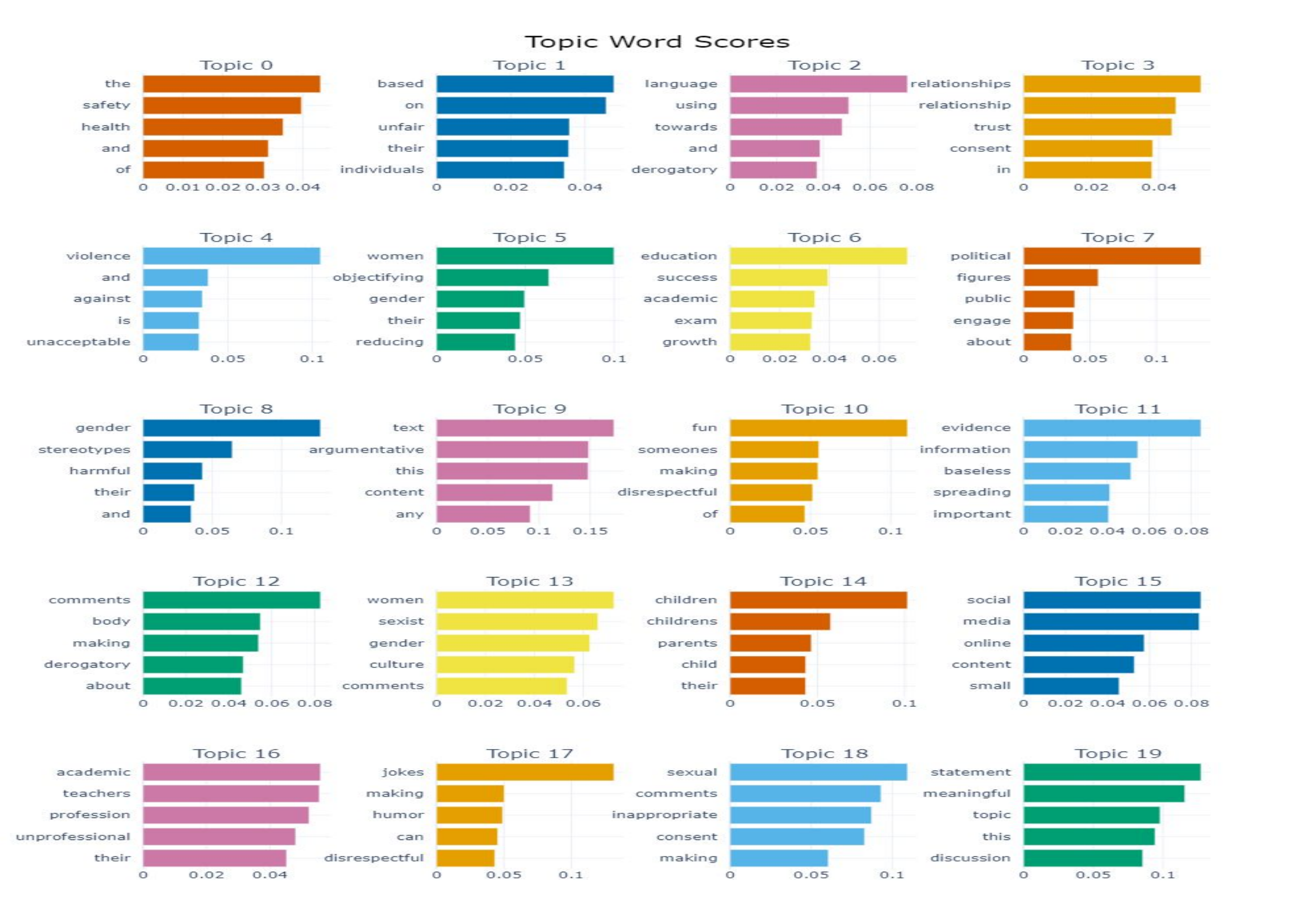}
	\caption {Topic of Human generated intervention.}
	\label{fig:human}
\end{figure*}

\begin{figure*}[hbt]
	\centering
 \includegraphics[height = 17cm, width = 17cm]{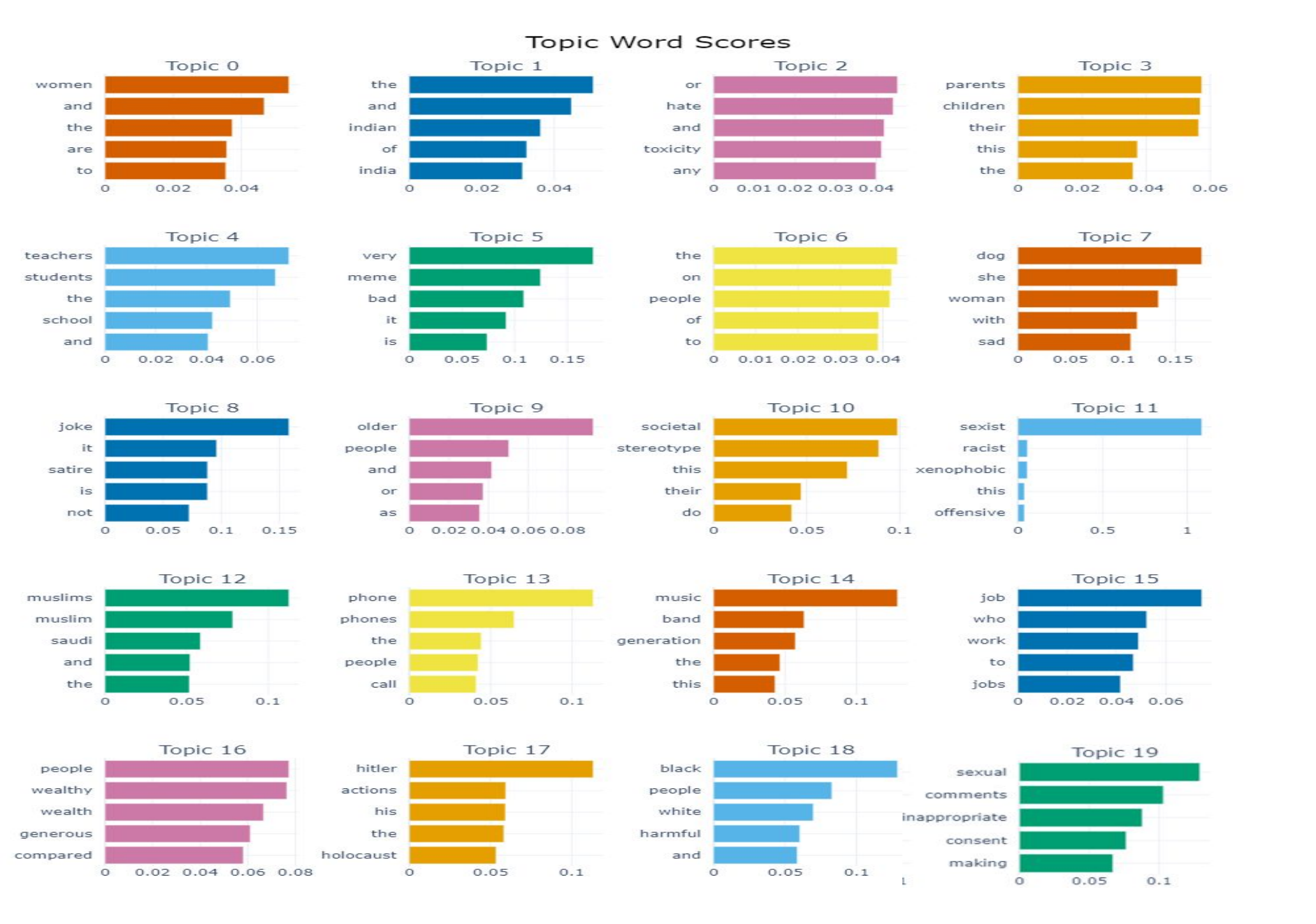}
	\caption {Topic of FLAN generated intervention.}
	\label{fig:flan}
\end{figure*}

\begin{figure*}[hbt]
	\centering
 \includegraphics[height = 17cm, width = 17cm]{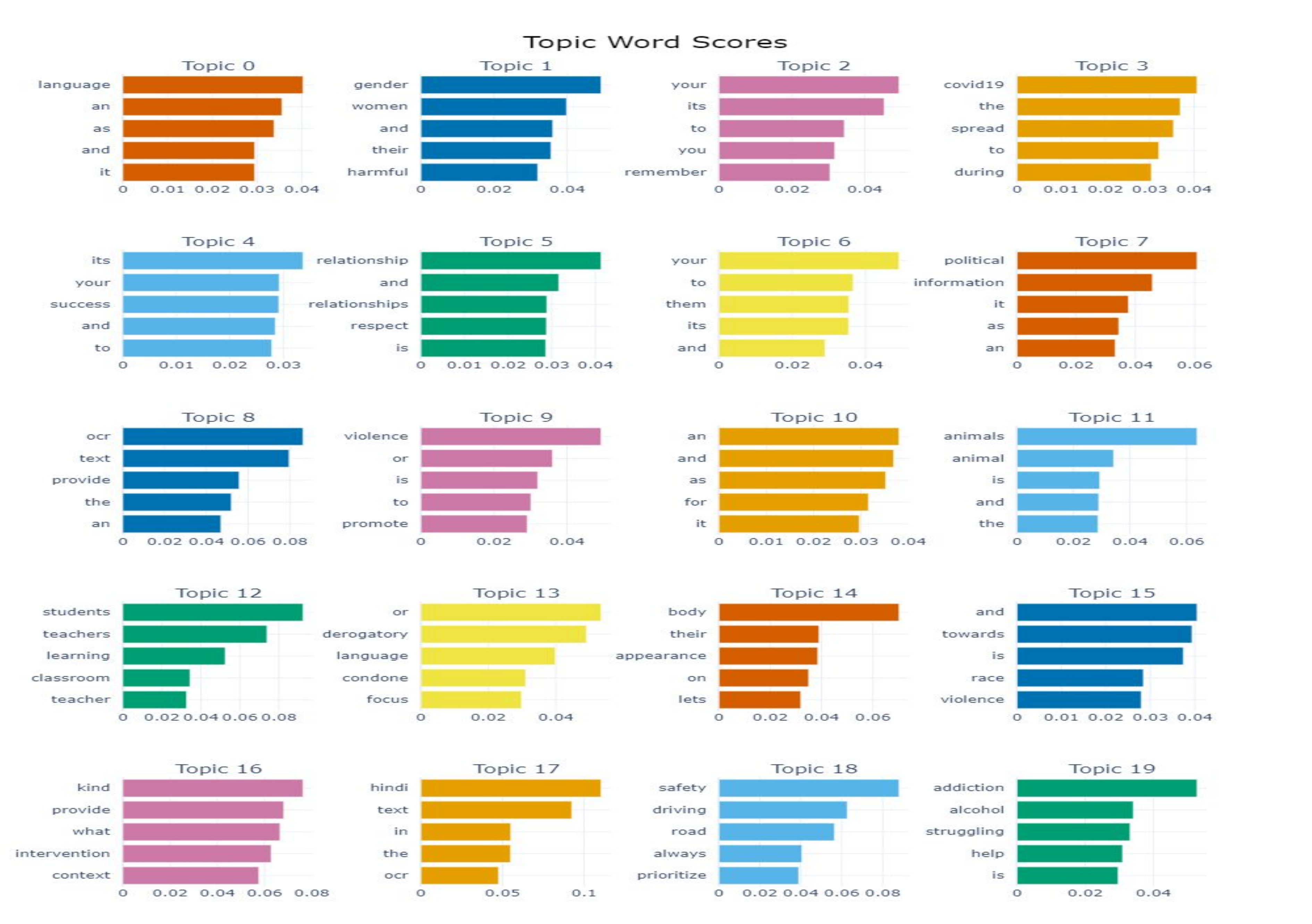}
	\caption {Topic of GPT3.5-Turbo generated intervention.}
	\label{fig:GPT3.5-Turbo}
\end{figure*}

We conducted a topical analysis of human-annotated interventions, FLAN-generated interventions, and GPT3.5-Turbo-generated interventions to assess the correspondence between interventions generated by Language Model (LLM) systems and those created by humans as shown in Figure~\ref{fig:human},~\ref{fig:flan}, and ~\ref{fig:GPT3.5-Turbo}. To perform this analysis, we utilized BERTopic~\cite{grootendorst2022bertopic}, a neural topic modeling approach that incorporates a class-based TF-IDF procedure. By projecting the top 20 topics, we gained insights into the thematic content of the interventions.

Figure~\ref{fig:human} presents the topics covered in the human-annotated interventions, encompassing a wide range of areas such as safety, health, fairness, trust, relationships, violence, gender, education, politics, misinformation, derogatory comments, and sexual abuse. In contrast, FLAN-generated interventions (Figure~\ref{fig:flan}) predominantly address topics related to gender, nationality, education, societal bias and stereotypes, race, violence, derogatory remarks, and sexual abuse. Additionally, Figure~\ref{fig:GPT3.5-Turbo} displays the topics addressed by GPT3.5-Turbo-generated interventions, which include gender, misinformation, relationships, politics, violence, education, physical appearance, safety, offensive language, and addiction.

Overall, there is a notable degree of alignment between the topics covered in the interventions generated by LLMs, such as FLAN and GPT3.5-Turbo, and those found in the human-annotated interventions.

\textbf{Qualitative Analysis: }
We also conducted a qualitative analysis of the generated interventions for the best-performing model on the ICMM dataset. Table~\ref{app:qual_appen} presents examples of interventions generated by both FLAN-T5 and GPT3.5-Turbo, utilizing MemeGuard and only OCR text. In the first example (A) shown in Table~\ref{fig:qual_appen}, both GPT3.5-Turbo and FLAN-T5 misinterpret the underlying gender association of the word "dishwasher" depicted in the meme with Only OCR Text. However, incorporating MeMeGuard it captures the associated gender roles and underlying association of "dishwasher" with women. Moving on to example (B) depicted in Table~\ref{fig:qual_appen}, GPT3.5-Turbo with MemeGuard adequately captures the hateful implication of the term "illegals" with respect to "immigration". However, FLAN-T5 with MemeGuard fails to capture the toxic element present in the meme, instead emphasizing more on the political policies of Donald Trump. Both GPT3.5-Turbo and FLAN-T5 with only OCR text struggle to adequately capture the implicit hateful implications. However, GPT3.5-Turbo appears to be more persuasive in both cases. Through this qualitative analysis, it becomes evident that the MemeGuard framework plays a crucial role in enabling the models to better understand and address the implicit hateful elements in the memes. GPT3.5-Turbo, in particular, demonstrates a stronger ability to generate interventions that align with the desired criteria of fluency, adequacy, persuasiveness, and informativeness.

\begin{table*}[]
\scalebox{0.49}{%
\begin{tabular}{|cc|c|c|}
\hline
\multicolumn{2}{|c|}{\textbf{Hateful Memes}} & (A) \begin{minipage}{0.78\textwidth}
\centering
      \includegraphics[width=\linewidth, height=60mm]{women_368.png}
    \end{minipage}                                & (B) \begin{minipage}{0.75\textwidth}
\centering
      \includegraphics[width=\linewidth, height=60mm]{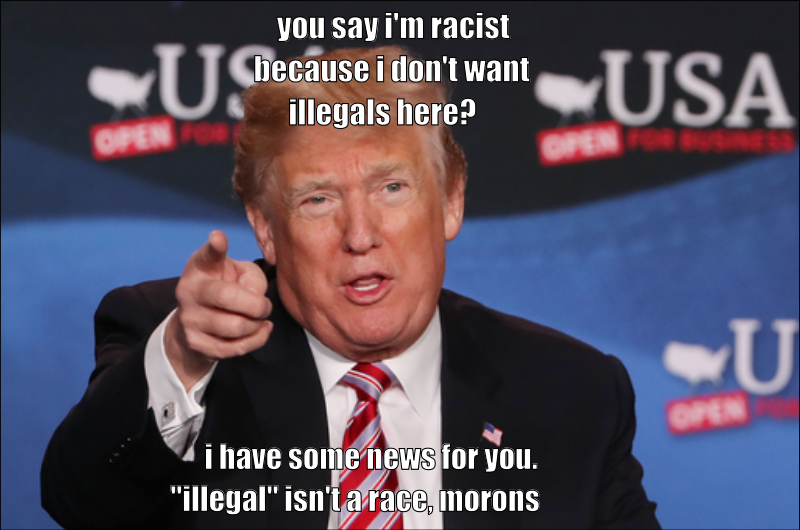}
    \end{minipage}                                                     \\ \hline
\multicolumn{2}{|c|}{\textbf{OCR Text}}      &\begin{tabular}[c]{@{}l@{}} a real man loads the dishwasher every night  \end{tabular} &  \begin{tabular}[c]{@{}l@{}}You say I am racist because I don’t want illegals here?\\ I have some news for you. "illegal" isn't a race, morons  \end{tabular} \\ \hline 
\multicolumn{1}{|c|}{}  & \textbf{OCR Text}  & \begin{tabular}[c]{@{}l@{}}Consistency in loading the dishwasher daily maintains cleanliness, but it's\\ equally vital to consider individual circumstances and needs in setting routines.\end{tabular} & I have some news for you. "illegal" isn't a race, morons. \\ \cline{2-4} 
\multicolumn{1}{|c|}{\multirow{-2}{*}[3ex]{{\textbf{FLAN-T5}}}} &
  \textbf{MemeGuard} &
  \begin{tabular}[c]{@{}l@{}} 
This meme reinforces harmful stereotypes about men and their roles in the \\household. A real man is not defined by their ability or willingness to load\\ the dishwasher; it is a household chore equally shared between partners. \\Interventions could include addressing the underlying stereotypes and \\promoting shared responsibility and respect in domestic relationships.\end{tabular} &
  \begin{tabular}[c]{@{}l@{}}This meme shows a bias against Donald Trump and his policies, but it also\\ conveys a general bias against any political figure who makes mistakes and\\ needs to be held accountable for their actions.\end{tabular} \\ \hline
\multicolumn{1}{|c|}{} &
 \textbf{OCR Text} &
  \begin{tabular}[c]{@{}l@{}} It is important to recognize that tasks such as loading the dishwasher should\\ not have gender roles attached to them. Both men and women should be\\ responsible for household chores, and it should not be labeled as a "real man"\\ task. Rather, all individuals living together in a household could take turns in \\completing these tasks, creating a more equal and fair division of labor. Gender \\stereotypes limit our potential and should be challenged.\end{tabular} &
  \begin{tabular}[c]{@{}l@{}}It is important to acknowledge that the statement "illegal isn't a race" is true,\\ however, it does not excuse any form of discriminatory behavior towards\\ individuals. As much as possible, discussions and debates should stick to\\ facts and avoid using language that may be offensive or derogatory. \end{tabular} \\ \cline{2-4} 
\multicolumn{1}{|c|}{\multirow{-2}{*}[10ex]{{\textbf{GPT3.5-Turbo}}}} &
  \textbf{MemeGuard} &
  \begin{tabular}[c]{@{}l@{}}
  Posting memes that reinforce gender stereotypes and biases can be harmful and \\contribute to a toxic culture. Let's challenge stereotypes and biases, and encourage\\ others to do the same. The next time you come across a meme like this, take a \\moment to reflect on its message and consider the harm it may cause. Let's use\\ social media as a tool to spread positivity and inclusivity, and promote a more \\accepting and equitable world.
  
  \end{tabular} &
  \begin{tabular}[c]{@{}l@{}}It is important to understand that using the term "illegals" to refer to\\ undocumented immigrants is dehumanizing and can perpetuate harmful\\ stereotypes. It is possible to have concerns about immigration without\\ resorting to language that is offensive and divisive. Instead, let's focus on\\ having respectful and productive conversations about immigration policy that\\ prioritize the safety and well-being of all individuals, regardless of their\\ immigration status.\end{tabular} \\ \hline 
\multicolumn{2}{|c|}{\textbf{Annotated Interventions}   } &
  \begin{tabular}[c]{@{}l@{}}Defining rigid gender roles that prescribe certain behaviors for "real men" and \\dehumanizing women by reducing them to mere sexual objects or limiting their\\ worth to domestic chores perpetuates harmful stereotypes.Promoting equality in \\household chores can lead to healthier and more balanced relationships, \\benefiting both partners.\end{tabular} &
  \begin{tabular}[c]{@{}l@{}}Using the term "illegals" to generalize and discriminate against non-citizens\\ is xenophobic and perpetuates harmful stereotypes.We should strive to\\ create a society that values diversity and promotes inclusivity, where\\ individuals are not judged based on their immigration status or ethnicity.\end{tabular} \\ \hline 
\end{tabular}
 }
\caption{Sample interventions  generated by best two \textit{MemeGuard} models and their corresponding baselines}
\label{app:qual_appen}
\vspace{-1mm}
\end{table*}

\end{document}